\documentclass{svproc}

\usepackage[utf8]{inputenc}
\usepackage[normalem]{ulem}
\usepackage{amsmath}
\usepackage{amssymb}
\usepackage{graphicx}
\usepackage{appendix}
\usepackage{cite}
\usepackage[table]{xcolor}
\usepackage[ruled,vlined,noend]{algorithm2e}
\usepackage{todonotes}
\usepackage{multicol}
\usepackage{footmisc}
\usepackage{wrapfig}
\usepackage{tikz}
\usetikzlibrary{shapes,positioning,decorations.pathreplacing}
\newtheorem{assumption}{Assumption}

\definecolor{matlabBlue}{rgb}{0, 0.4470, 0.7410}
\definecolor{matlabOrange}{rgb}{0.8500, 0.3250, 0.0980}
\definecolor{matlabYellow}{rgb}{0.9290, 0.6940, 0.1250}
\definecolor{matlabPurple}{rgb}{0.4940, 0.1840, 0.5560}
\definecolor{matlabGreen}{rgb}{0.4660, 0.6740, 0.1880}
\definecolor{matlabSkyBlue}{rgb}{0.3010, 0.7450, 0.9330}
\definecolor{matlabRed}{rgb}{0.6350, 0.0780, 0.1840}

\newcommand{\demo}{\xi^\dagger}

\newcommand{\X}{\mathbb{R}^2}

\newcommand{\env}{O}
\newcommand{\envset}{\mathbb{O}}
\newcommand{\knownregion}[1][t]{R(#1)}

\newcommand{\candtraj}{\xi_{\textrm{cand}}}
\newcommand{\spt}{\alpha}
\newcommand{\SPT}{\mathbb{A}}
\newcommand{\cell}{\lambda}

\newcommand{\Cell}{\Lambda}

\newcommand{\class}{\omega}
\newcommand{\proj}[1]{#1}
\newcommand{\Class}{\Omega}
\newcommand{\startCells}[1]{\Cell^{\textrm{start},#1}_{\textrm{full}}}
\newcommand{\safeset}{\mathcal{S}}
\newcommand{\fullCell}{\Cell_{\textrm{full}}}
\newcommand{\graphpath}{\sigma}
\newcommand{\graphpathvar}{\gamma}
\newcommand{\Graphpath}{\Sigma}

\newcommand{\node}{\eta}

\newcommand{\obsseg}{\zeta}
\newcommand{\numseg}{K_i}
\newcommand{\seg}{\nu}
\newcommand{\joinpt}{\beta}

\newcommand{\length}[1]{|#1|}

\newcommand{\nodecell}[1]{#1^\cell}
\newcommand{\nodesp}[1]{#1^\spt}

\newcommand{\numrays}{K^r_{\kappa,l}}

\newcommand{\maxassign}{I_{\textrm{assign}}}
\newcommand{\maxgrad}{I_{\textrm{grad}}}
\newcommand{\maxinitial}{I_{\textrm{initial}}}
\newcommand{\assign}{\mathcal{A}}
\newcommand{\param}{\theta}
\newcommand{\grad}{\mu}
\newcommand{\suchthat}{\enspace \text{s.t.} \enspace}
\newcommand{\plan}[1]{\tilde{#1}}
\newcommand{\adj}[3]{\mathcal{P}_{#3}(#1,#2)}
\newcommand{\occ}[3]{\textrm{PO}_{#3}(#1,#2)}
\newcommand{\fullyocc}[3]{\textrm{FO}_{#3}(#1,#2)}

\newcommand{\chainable}[2]{\chi(#1,#2)}
\newcommand{\classobs}[1]{\mathcal{O}(#1)}
\newcommand{\segdisc}{\Delta}
\newcommand{\solnenv}{\envset_{\textrm{soln}}}

\newcommand{\classid}{\mathbf{\Class}_{\textrm{id}}}

\title{Inferring Obstacles and Path Validity from Visibility-Constrained Demonstrations \vspace{-10pt}}
\titlerunning{Inferring Obstacles from Demonstrations}
\author{Craig Knuth, Glen Chou, Necmiye Ozay, Dmitry Berenson%
\thanks{\footnotesize{This research was supported in part by NSF grants IIS-1750489 and ECCS-1553873, ONR grants N00014-17-1-2050 and N00014-18-1-2501, and a National Defense Science and Engineering Graduate (NDSEG) fellowship.}}
}
\institute{University of Michigan, Ann Arbor, MI 48109, USA \\
\texttt{\{cknuth, gchou, necmiye, dmitryb\}@umich.edu}
}

\authorrunning{C. Knuth et al.}

\date{April 2019}

\begin{document}

\maketitle

\vspace{-20pt}
\begin{abstract} Many methods in learning from demonstration assume that the demonstrator has knowledge of the full environment. However, in many scenarios, a demonstrator only sees part of the environment and they continuously replan as they gather information. To plan new paths or to reconstruct the environment, we must consider the visibility constraints and replanning process of the demonstrator, which, to our knowledge, has not been done in previous work. We consider the problem of inferring obstacle configurations in a 2D environment from demonstrated paths for a point robot that is capable of seeing in any direction but not through obstacles. Given a set of \textit{survey points}, which describe where the demonstrator obtains new information, and a candidate path, we construct a Constraint Satisfaction Problem (CSP) on a cell decomposition of the environment. We parameterize a set of obstacles corresponding to an assignment from the CSP and sample from the set to find valid environments. We show that there is a probabilistically-complete, yet not entirely tractable, algorithm that can guarantee novel paths in the space are unsafe or possibly safe. We also present an incomplete, but empirically-successful, heuristic-guided algorithm that we apply in our experiments to 1) planning novel paths and 2) recovering a probabilistic representation of the environment.
\end{abstract}

\vspace{-30pt}

\section{Introduction}

\vspace{-8pt}

In the paradigm of learning from demonstration, the assumption is often made that the demonstrator is acting (near) optimally. However, in many real world scenarios, a demonstrator only has partial information about the environment and they continuously replan as they gather information. In the context of 2D navigation, such as a robot scouting an underwater wreck or a medical assistance team navigating an unknown environment, demonstrators are able to see obstacles only when they enter the field of view and plan a path to goal with only partial knowledge of the environment, i.e. parts of the environment that have not been occluded. If we cannot access the internal map of the demonstrator, such as an occupancy map of a robot performing SLAM or the memory of a human being, we may want to reconstruct the environment using the demonstration.

\begin{figure}[t]
        \vspace{-10pt}
        \centering
        \includegraphics[width=\textwidth]{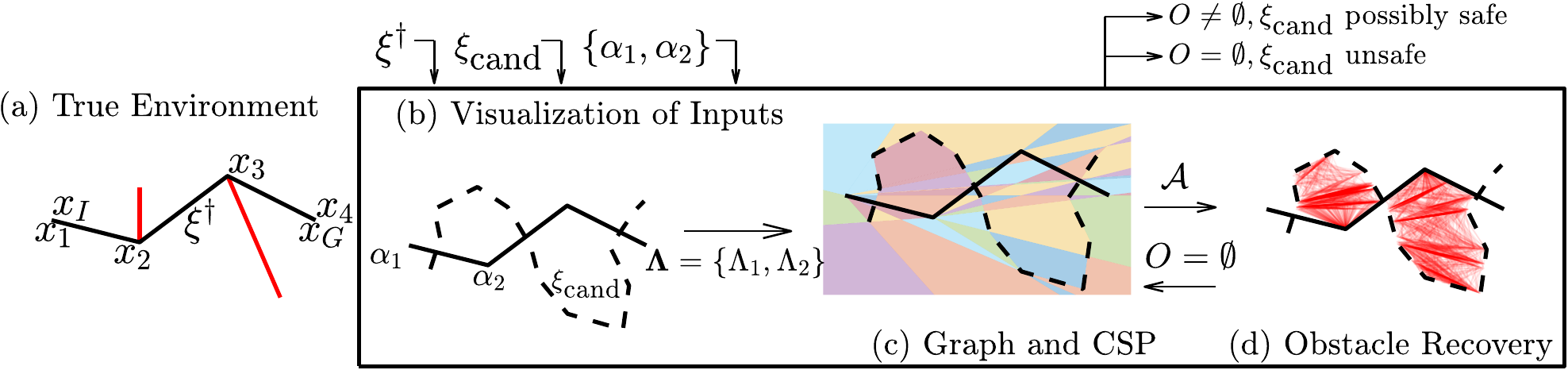}
        \vspace{-22pt}
        \caption{\footnotesize{An overview of the method (boxed in black). Given a demonstration $\demo$ and survey points $\SPT = \{\spt_1, \spt_2\}$, we test the validity of a candidate path (dashed black) by (b) decomposing the space into cells $\mathbf{\Cell}$ encoding visibility constraints, then sample obstacles $O$ from a set of obstacles $\assign$ produced by the CSP. If we find a valid obstacle configuration in $\assign$, the candidate is possibly safe. If not, then we query for a new $\assign$. In (d), we show many sample obstacles found by our method.}} 
        \label{fig:method_overview}
\end{figure}

Reconstructing the environment when considering visibility constraints can be challenging for a number of reasons. The demonstration may not be globally optimal, or even locally optimal
%, for the full-observation planning problem, 
as the robot may make suboptimal decisions due to the lack of global environment knowledge or backtrack upon gaining information. Thus, constraint-learning methods relying on global \cite{Chou2018LearningConstraints, Chou2019CoRL} or local optimality \cite{Chou2020RAL} of the demonstrations cannot be applied. The demonstrator is constantly obtaining more information about the environment and therefore plans paths not only based on what they currently see but also what they have already seen. The memory and vision model of the demonstrator means that obstacles in the world are all potentially intertwined; two obstacles may be individually consistent with our assumptions on the demonstration, but together are not. Furthermore, the constraint-learning problem is ill-posed \cite{Chou2018LearningConstraints} as many environment configurations can be consistent. Thus accounting for visibility constraints introduces unique requirements and challenges for environment reconstruction from demonstration. 
%To our knowledge, there is no existing environment reconstruction method that accounts for these kinds of constraints.

We utilize information from a learned environment configuration in two ways. First, in the context of planning, we determine if a novel robot path (for a new start and goal) is certainly unsafe or possibly safe.
% could include paragraph on algorithms that are possible in evaluating candidate safety
%We can also extend this approach to generate many novel paths so that we can choose from a set of initial plans.
Second, we probabilistically reconstruct an environment, for instance from the demonstrated path of a scouting robot, by generating many possible environment configurations with an associated probability measure. 
%Then, we can use this probabilistic representation of the environment for other applications such as planning or environment identification.
This method may be useful in scenarios where we do not have a map (such as a private home or disaster environment) but we do have demonstrations from agents such as medical professionals or first-responders.

Our learning approach is composed of the following steps. Given a demonstration or several sequential demonstrations, a set of survey points (points
from which the demonstrator sees the environment), and a novel candidate robot path, we perform a Polar Cell Decomposition (PCD) that captures occlusions to a particular survey point. Combining cell decompositions, we construct a graph that captures configurations of occluding obstacles. On this graph, we formulate a CSP 
%that ensures obstacles do not occlude themselves or one another with respect to the survey points of the demonstration, thus ensuring the demonstrator sees relevant obstacles at relevant times.
such that each assignment denotes a set of possible obstacles. We parameterize and sample from the set of obstacles
%and perform gradient ascent 
to find a valid environment configuration (see Fig. \ref{fig:method_overview}). In certain cases, we can assert no valid environment exists and conclude the novel candidate robot path is unsafe. To the best of our knowledge, this is the first paper to address the problem of reconstructing an environment from visibility-constrained demonstrations.

The specific contributions of this work are 
\textit{1)} two algorithms for generating possible 2D environment configurations subject to visibility-constrained demonstrations; 
\textit{2)} a theoretical analysis of the method; 
\textit{3)} an extension to planning novel paths in the demonstrator's environment; 
\textit{4)} an extension for constructing a probabilistic representation of the environment; and 
\textit{5)} experimental results of running the algorithm on a scouting robot and in-home example.

\vspace{-15pt}
\section{Related Work}
\vspace{-6pt}

Our method is closely related to the literature in inverse optimal control/inverse reinforcement learning (IOC/IRL) \cite{Abbeel:2004:AL_IRL, ng_irl, ARGALL_survey_lfd,kalman,Ratliff2006MaximumMP}. These methods aim to learn a cost function from demonstrations which when optimized, generalize behavior present in the demonstrations. However, by only searching for cost functions, IOC/IRL can be poorly suited for representing richer classes of task specifications, i.e. those requiring satisfaction of hard constraints. In this work, we take steps to bridge this gap by inferring obstacle constraints, assuming that the demonstrator is acting optimally in a visibility-constrained, receding horizon fashion.

Our work is also related to constraint-learning. Much work has focused on learning local trajectory-based constraints \cite{dmitry, anca, lfdc1, lfdc2, lfdc3, lfdc4}, i.e. task-specific constraints which need not hold globally over all trajectories, while other methods \cite{vijayakumar, shah} focus on learning geometric constraints. The closest methods in the literature are \cite{Chou2018LearningConstraints, Chou2019CoRL} and \cite{Chou2020RAL}, which learn safe and unsafe regions of the state space from globally and locally optimal expert demonstrations, respectively, assuming that the demonstrator has full knowledge of the environment \textit{a priori}. This paper removes this assumption, instead focusing on demonstrators that plan optimally while incrementally gathering information about the environment. Removing this assumption fundamentally changes the nature of the problem, requiring a radically different algorithm than previous work.

\vspace{-10pt}
\section{Preliminaries}
\vspace{-5pt}

\subsection{Demonstrator's Problem}
\vspace{-3pt}

We consider a demonstrator acting as a point robot with purely kinematic constraints in $\X$. They start at position $x_I$ and attempt to find a collision-free path to the goal $x_G$ in the presence of static obstacles (or environment) $\env \subset \X$ which is incrementally revealed via exploration resulting in a demonstration $\demo$.

Given two points $p$ and $q$, $q$ is visible from $p$ if and only if the open-ended line segment connecting them, denoted as $(\overline{pq} \setminus \{p,q\})$ does not intersect $\env$. Note $q$ may lie in $O$. On the demonstrated robot path $\demo$ at time $t$, the demonstrator sees a region $V(t) \doteq \{q \,|\,q\,\text{is visible from}\, \xi(t)\}$ (a 360$^\circ$ view with no uncertainty). This sensor model applies to a 360$^\circ$ camera or an omni-directional lidar, which are frequently used on mobile robots. As the demonstrator moves about the environment, they expand the known region $\knownregion \doteq \bigcup_\tau V(\tau)$, $0 \leq \tau \leq t$ and the known obstacles $O(t) \doteq \knownregion \cap \env$.  %Such a sensor model is reasonable for We justify the sensing model of the demonstrator by noting the prevalence of robotic systems that have a 360$^\circ$ view (such as a rover).

We assume the demonstrator follows a strategy of planning an optimal path under the known obstacles $O(t)$, following that path a small amount, and then replanning. This strategy is formalized below. Let $\Xi \doteq \{\xi \,|\, \xi : [0, 1] \to \X\}$ denote the set of paths from any start to any goal, $c : \Xi \rightarrow \mathbb{R}$ be a cost function, and $a$ and $b$ are two points in $\mathbb{R}^2$.
%The problem of finding an optimal robot path $\plan\xi$ between two points $a$ and $b$ with known obstacles $O(t)$ and cost function $c : \Xi \rightarrow \mathbb{R}$ is shown in Problem \ref{prob:partially_optimal}:

\vspace{-3pt}
\begin{problem}[Demonstrator's Planning Problem \footnote{\scriptsize A minimum may not exist, but discussion is excluded for brevity. See proof of Thm. \ref{th:exit_point}}]\label{prob:partially_optimal}
\vspace{-6pt}
\begin{equation}\label{eq:partially_optimal}
    \begin{split}
        \begin{aligned}
            \plan\xi \enspace = \quad& \text{arg min}_{\xi}
            & & c(\xi) \\
            & \text{subject to}
            & & \xi(0) = a, \enspace \xi(1) = b, \enspace \xi(\tau) \,\notin\, O(t) \enspace \forall \tau \in [0, 1] \\
        \end{aligned}
    \end{split}
\end{equation}
\vspace{-15pt}
\end{problem}

We use the symbol \textasciitilde $\,$ (e.g. $\plan\xi$) to denote a path planned with respect to a fixed obstacle configuration to distinguish it from the executed demonstration $\demo$. We consider cost to be the total path length. We emphasize that $\plan\xi$ may not coincide with $\demo$ as more of the environment is revealed. For example, the path $\xi_{inf}^1$ in Figure \ref{fig:inf_alt} 
%(which aligns with the solution to Prob. \ref{prob:partially_optimal} starting at $x_1$)
intersects with an obstacle that is not visible from $x_1$ and must be adjusted at $x_2$. However, from the following problem we know that the demonstration $\demo$ and plan $\plan\xi$ will align for at least a small segment which will allow us to verify a demonstration with respect to any obstacle $O$.
%, i.e. for a path $\xi$ composed of straight line segments between vertices $\{y_i\}_{i=1..k}$, $c(\xi) = \sum_{i=1}^{k-1} l(y_iy_{i+1})$ where $l(\cdot)$ is the length of the segment. % to save space
%The demonstrator assumes that all unknown space is free, plans a path $\plan\xi$ to goal, follows $\plan\xi$ for some nonzero length, and then replans a new $\plan\xi$.

\vspace{-5pt}
\begin{problem}[Demonstrator's Strategy]\label{prob:arrive_at_goal}
Find $\demo:[0,1] \rightarrow \X$ such that $\demo(0) = x_I$, $\demo(1) = x_G$ and is generated with the following strategy. At any time $t$ with the demonstrator at position $p$, the demonstrator follows $\plan\xi$, the solution to Prob. \ref{prob:partially_optimal} for start $p = \demo(t)$, goal $g$, and known obstacles $O(t)$, i.e. $\exists \delta > 0, u > 0$ such that $\demo(t + \frac{\tau}{u}) = \plan\xi(\tau)$ $\forall \tau \in [0, \delta)$.
\end{problem}
\vspace{-5pt}

The final condition enforces that the demonstrator follows $\plan\xi$ at each time $t$ for some nonzero length. We introduce the scaling factor $u$ for time-scaling purposes to ensure $\demo$ is defined only on $[0,1]$. We highlight that even though the demonstrator plans optimally with respect to $O(t)$, we do not have access to the entire planned path because the demonstrator's plan may change as $O(t)$ changes. Thus methods like \cite{Chou2018LearningConstraints, Chou2019CoRL} are not applicable.

\vspace{-7.5pt}
\subsection{Obstacles and Obstacle Vertices}

\vspace{-4pt}
\begin{assumption}[Line Segment Obstacles]\label{assume:line_segments}
We assume obstacles are curves consisting of only straight line segments.
\end{assumption}
\vspace{-5pt}

A consequence of Assumption \ref{assume:line_segments} is that the demonstration can also be decomposed into line segments connecting vertices $\{x_i\}_{i=1}^{n+2} = \{\demo(t_i)\}_{i=1}^{n+2}$ where $n+2$ is the number of vertices on the demonstration, and each intermediate vertex $\{x_i\}_{i=2}^{n+1}$ is coincident with an obstacle vertex. We can consider this as an extension to the well-known theorem that the shortest path between any two points in a 2D environment with polygonal obstacles is composed of straight line segments with intermediate vertices at obstacle vertices \cite[Visibility Graphs]{CompGeom}. There is a subtle distinction here in that the planned path $\plan\xi$ will always have vertices at obstacle vertices, but we show in the following theorem that the executed robot path $\demo$ (during which the known obstacles and planned paths may change) will also have vertices close to obstacles. 
\vspace{-3pt}
\begin{theorem}\label{th:exit_point}
At time $t$, suppose $\plan\xi$ solves Prob. \ref{prob:partially_optimal}, $\plan\xi$ is not a straight line to goal, and $x^* = \plan\xi(s)$ is the first point on $\plan\xi$ coincident with an obstacle vertex. Then, $\demo(t+\tau) = \plan\xi(\tau)$ for all $\tau \in [0, s]$. Therefore, the demonstrator will never deviate from a plan unless at an obstacle vertex.
\end{theorem}
\vspace{-5pt}

%An intuitive explanation of Theorem \ref{th:exit_point} is that the demonstrator will never deviate from a plan until they reach the next obstacle vertex. 
%Therefore, each intermediate vertex is coincident with an obstacle vertex. % space saver
For sake of brevity, we present all proofs in the appendix. 
%As all curves can be closely approximated with line segments, we do not consider Assumption \ref{assume:line_segments} to be restrictive. 
Assumption \ref{assume:line_segments} guarantees our method will produce sound results, but two more assumptions are needed to ensure the completeness of our method.

\vspace{-6pt}
\begin{assumption}[Non-intersecting Environment]\label{assume:non_spoked}
We assume the environment does not contain obstacles which intersect with themselves or one another.
\end{assumption}

\vspace{-13pt}
\begin{assumption}[$n$ Obstacle Curves]\label{assume:n_curves}
We restrict the environment to contain at most $n$ obstacle curves each with an endpoint at some $x_i$, $i \in \{2, \ldots, n+1\}$. Consequently, there are no ``free-standing" obstacles, i.e. obstacles that do not intersect with an intermediate vertex of the demonstration.
\end{assumption}
\vspace{-8pt}

%We note that even though we cannot learned enclosed obstacles, there are many environments corresponding to an environment with an enclosed obstacle that produce the same demonstration. We demonstrate this in the Results section. We leave reconstruction of spoked, enclosed, and free-standing obstacles to future work.

%With the assumptions on the environment and demonstration stated, we make a note on learnability. First, at the third to last vertex of the demonstration, the plan of the demonstrator coincides with the rest of the demonstration. Therefore, we can only learn the subset of the environment visible at the third to last vertex, i.e. $R(t_n)$. % in depth commentary on the learnability of the environment

\vspace{-12pt}
\subsection{Verifying Environment Configurations}\label{sec:verifying_conf}
\vspace{-3pt}

\begin{wrapfigure}{l}{0.37\textwidth}
    \centering
    \vspace{-22pt}
    \includegraphics[width=0.37\textwidth]{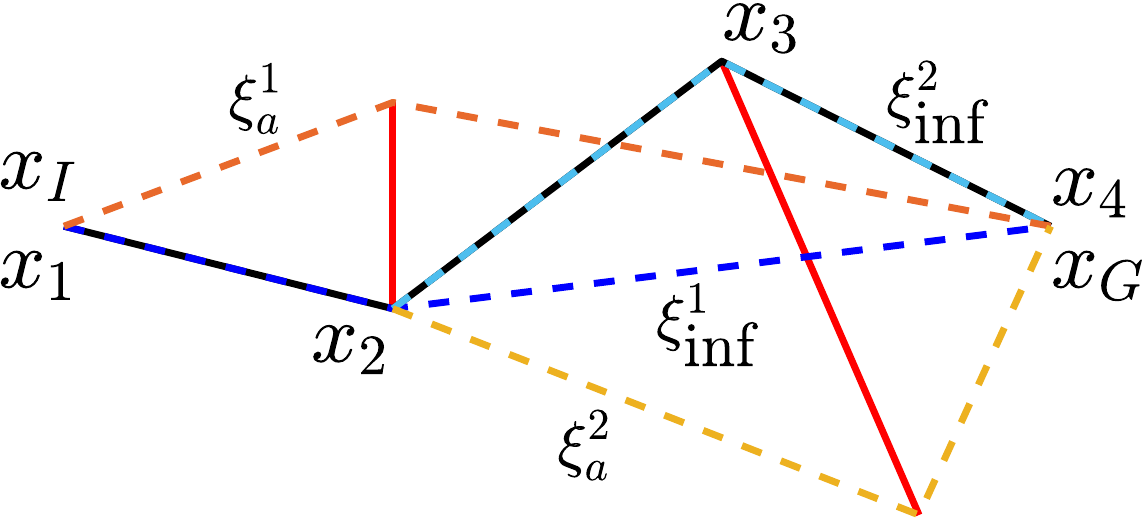}
    \vspace{-22pt}
    \caption{\footnotesize{Inferred $\xi^i_{\textrm{inf}}$ (blue, light blue) and alternative $\xi^i_a$ (orange, yellow) robot paths from vertices $x_1$ and $x_2$. Demo is solid black and obstacles are red. Note that the first line segment of each $\xi^i_{\textrm{inf}}$ aligns with the next line segment of the demonstration. Here $\plan\xi^i$ is precisely $\xi^i_{\textrm{inf}}$.}}
    \label{fig:inf_alt}
    \vspace{0pt}
\end{wrapfigure}

Thm. \ref{th:exit_point} informs how we can verify that $\demo$ is the solution to Prob. \ref{prob:arrive_at_goal} for a given environment configuration $\env$. If all $\plan\xi^i$'s,  defined to be the solutions to Prob. \ref{prob:partially_optimal} with start $x_i$ and goal $x_G$ 
for $1 \leq i \leq n$%
, align with the straight line from $x_i$ to $x_{i+1}$ then $\demo$ is the solution to Prob. \ref{prob:arrive_at_goal} (see Fig. \ref{fig:inf_alt}). We can verify an environment by only considering plans originating at the first $n$ vertices by Thm. \ref{th:exit_point}.

%We will adopt the following strategy to verify an environment configuration $O$. 
For each vertex $x_i$ of $\demo$ with known obstacles $O(t_i)$, $i \in \{1, \ldots, n\}$, let $\mathbb{H}^i$ be the set of homology classes of robot paths between $x_i$ and $x_G$, excluding homology classes that wind around obstacles at least once. For further treatment on homology classes in the context of finding robot paths see \cite{Bhattacharya2010SearchBasedPP}. For each class $H^i \in \mathbb{H}^i$ we calculate the shortest robot path. At least one of these robot paths (there may be multiple, say $N^i_{\textrm{inf}}$) will overlap with the next line segment of the demonstration. We define $\plan\xi^i_{\textrm{inf}}$ as the shortest of these paths, indicating this is the \textit{inferred robot path}. We say this robot path is inferred because it may or may not be the same as the robot path the demonstrator actually planned, depending on the true environment. The rest of the $N^i_a = |H^i| - N^i_{\textrm{inf}}$ robot paths that do not overlap with the next segment of the demonstration are the \textit{alternative robot paths} $\{\plan\xi^i_{a_k}\}_{k=1}^{N^i_a}$ which leads us to the next definition. Let $\plan\xi^i_a$ be the shortest of these alternatives.

\vspace{-5pt}
\begin{definition}[Consistency]\label{def:consistency}
If for some obstacle $O$, $c(\plan\xi^i_{\textrm{inf}}) \leq c(\plan\xi^i_a)$ for all $i \in \{1, 
\ldots, n\}$, we say the obstacle is \textit{consistent with the demonstration}.
\end{definition}
\vspace{-5pt}

If $O$ is consistent with a demonstration $\demo$, then $\demo$ solves Prob. \ref{prob:arrive_at_goal} for $O$. Determining consistency by examining plans only at the first $n$ vertices implies that we can only learn the portion of the environment visible by the n$^{th}$ vertex.
%Therefore, we restrict our search to obstacles that lie in $R(t_n)$.

\vspace{-12pt}
\section{Problem Statement}\label{sec:prob_statement}
\vspace{-8pt}

We assume we are given a finite set of survey points $\SPT \doteq \{\spt_1, \spt_2, \ldots, \spt_{N_\spt}\} = \{\demo(s_1), \ldots, \demo(s_{N_\spt})\}$ such that the known region at each time $s_j$ can be exactly reconstructed from only the visible region at each survey point, i.e. $R(s_j) = \bigcup_{s_k \leq s_j} V(s_k)$. We assume that $\SPT$ contains $\{x_1, \ldots, x_n\}$ (all points of the demonstration except the last two). Intuitively, these survey points describe the points at which the demonstrator takes in information useful for planning. Note the demonstrator continuously sees more of the environment whereas in reconstruction we have a finite number of survey points which will enable a discrete search.
%This enables our method to perform a discrete search over obstacles.

Given a demonstration $\demo$ and survey points $\SPT$, we evaluate the safety of a given candidate robot path $\candtraj$. We assume $\candtraj$ is safe and then seek to find a consistent environment $\env$. If we are able to find an environment then we can conclude the candidate is possibly safe otherwise it is certainly unsafe.
%as it is safe in some possible world. If not then the candidate is certainly unsafe.

\vspace{-5pt}
\begin{problem}\label{prob:cand_traj}
Given a demonstration $\demo$, survey points $\SPT$, and a candidate robot path $\candtraj$, find a consistent environment configuration $\env$ such that $\env \cap \candtraj = \emptyset$.
\end{problem}
\vspace{-5pt}

As an extension, we also generate a probabilistic reconstruction of the environment, potentially from multiple sequential demonstrations as in the case of the scouting robot. 
%For example, a robot commanded to visit several waypoints in an unknown environment would produce several sequential demonstrations, $\{\demo_k\}_{k=1}^{K}$ with $\demo_k(1) = \demo_{k+1}(0)$ for all $k \in \{1, \ldots, K-1\}$.
Let $\{\demo_k\}_{k=1}^{K}$ with $\demo_k(1) = \demo_{k+1}(0)$ for all $k \in \{1, \ldots, K-1\}$ be a set of \textit{sequential} demonstrations.

\vspace{-5pt}
\begin{problem}\label{prob:env_prob}
Given $K$ sequential demonstrations $\{\demo_k\}_{k=1}^{K}$ and a prior on obstacles, find environment configurations $\envset$ and a probability measure $\mathbb{P} : \envset \to [0, 1]$ such that $\mathbb{P}$ is consistent with the prior and for each $\env \in \envset$ with $\mathbb{P}(O) > 0$ and each start $\demo_k(0)$ and goal $\demo_k(1)$ in $k \in \{1, \ldots, K\}$, the solution to Prob. \ref{prob:arrive_at_goal} is $\demo_k$.
\end{problem}
\vspace{-5pt}

\vspace{-8pt}
\section{Method}
\vspace{-6pt}

We present an algorithm to find an environment configuration $O$ that solves Prob. \ref{prob:cand_traj} under Assumptions \ref{assume:line_segments} - \ref{assume:n_curves}.
%To find these environment configurations, we search for obstacles that do not intersect with $\candtraj$ in the environment as well as are visible to the demonstrator at appropriate survey points. 
In Sec. \ref{sec:method1}, we present a potentially intractable algorithm that is probabilistically complete which finds an environment configuration that solves Prob. \ref{prob:cand_traj} if one exists. In Sec. \ref{sec:method2} we present an incomplete but tractable variant.
%For instance, we want to be able to search for environment configurations where some obstacles are hidden until the demonstrator rounds a corner. 
First we give a few more definitions. Define the safe set $\safeset \doteq \{\demo(\tau) | \tau \in [0,1]\} \cup \{\candtraj(\tau) | \tau \in [0,1]\}$.

\vspace{0pt}
\begin{definition}[Full/Partial Occlusion]
A set $X \subset \X$ is \textit{fully occluded} from point $p$ by a set $Y$ if $Y \cap X = \emptyset$ and for all $q \in X$, $\overline{pq} \cap Y_1 \neq \emptyset$. Let $\fullyocc{X}{Y}{p}$ be a proposition that is true in this case and otherwise false. Furthermore, $X$ is \textit{partially occluded} from $p$ by $Y$ if there exists a point $q \in X$ such that $\overline{pq} \cap Y \neq \emptyset$. Let $\occ{X}{Y}{p}$ be defined similarly.
\end{definition}
%\vspace{-5pt}

%\vspace{-5pt}
%\begin{definition}[Partial Occlusion]
%A set $Y_2 \subset \X$ is \textit{partially occluded} from point $p$ by a set $Y_1$ if there exists a point $q \in Y_2$ such that $\overline{pq} \cap Y_1 \neq \emptyset$. Let $\occ{Y_2}{Y_1}{p}$ be a proposition that is true in this case and otherwise false.
%\end{definition}
%\vspace{-5pt}

%\vspace{-5pt}
%\begin{definition}[Assumed Safe Set]
%We define the assumed safe set as
%$\safeset = \{\demo(\tau) | \tau \in [0,1]\} \cup \{\candtraj(\tau) | \tau \in [0,1]\}$
%\end{definition}
%\vspace{-5pt}

\vspace{-12pt}
\subsection{Probabilistically Complete Obstacle Sampling}\label{sec:method1}
\vspace{-3pt}

Our method first determines the polar cell decomposition (PCD) from each survey point, combines the decompositions and survey points in a graph, formulates a constraint satisfaction problem (CSP) on the graph, and then samples obstacles corresponding to an assignment of the CSP.

\vspace{-10pt}
\subsubsection{PCD}\label{sec:pcd}

%A \textit{polar cell decomposition} of the plane  in the presence of lines $\{l_1, l_2, \ldots\}$ from the point $\alpha$ is a partition of $\mathbb{R}^2$ into \textit{cells} $\{\cell_0, \cell_1, \ldots\}$ such that interiors of cells are disjoint, 

\begin{wrapfigure}{r}{0.32\textwidth}
	\centering
	\vspace{0pt}
	\includegraphics[width=0.32\textwidth]{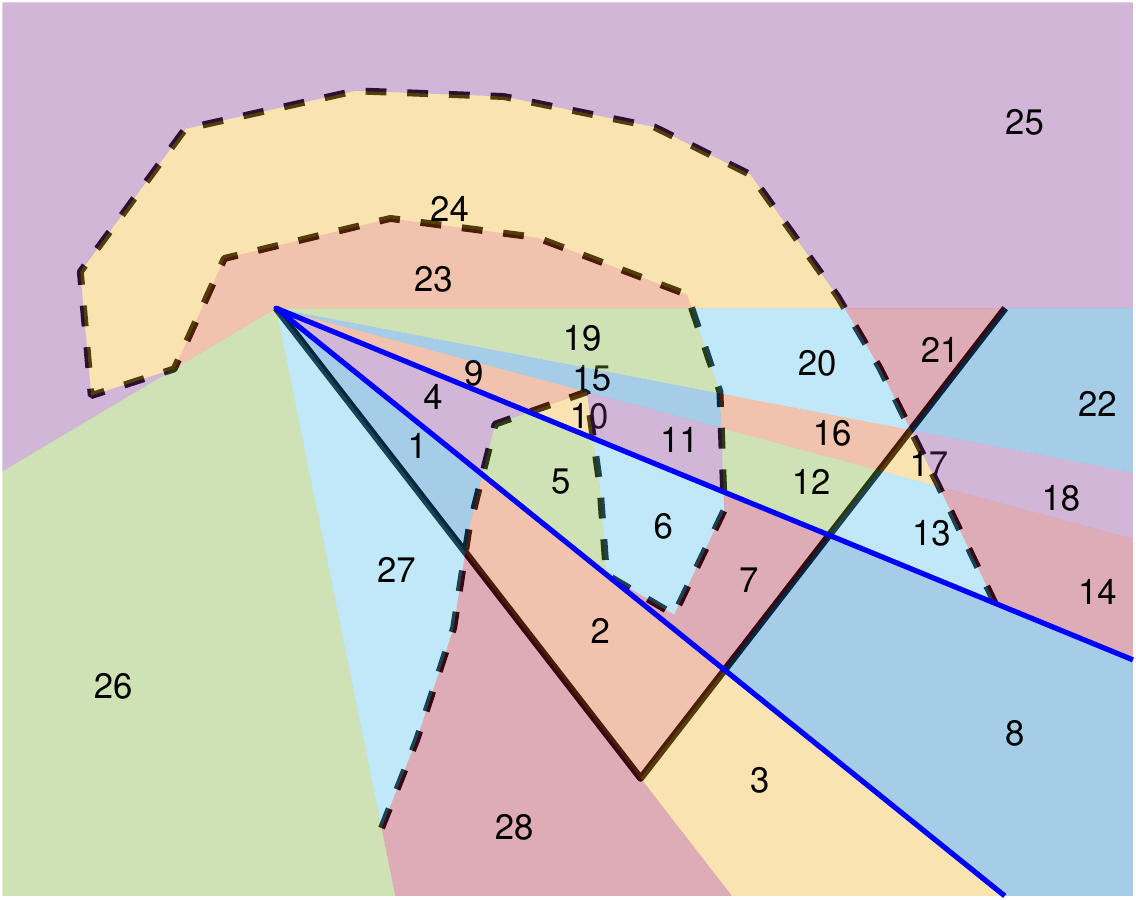}\vspace{-6pt}
	\caption{\footnotesize{A PCD with $\demo$ (solid black) and candidate robot path $\candtraj$ (dashed black). The cells outlined in blue all share the same bounding rays so nearer cells occlude further ones, e.g. cell 5 occludes cell 7. Cell sequence identification would give $\omega_1 = 2 \to 7 \to 12 \to 16 \to 20 \to 24$.}}
	\label{fig:pcd}
	\vspace{-5pt}
\end{wrapfigure}

We perform an exact cell decomposition of the space with respect to the demonstration $\demo$ and candidate robot path $\candtraj$ by sweeping rays from each survey point $\alpha_j$ resulting in a \textit{polar cell decomposition} $\Cell_{\spt_j}% = \{\cell_1^{\spt_i}, \ldots, \cell_{|\Cell_{\spt_i}|}^{\spt_i}\}%
$. Each cell $\cell \subset \mathbb{R}^2$ is closed and shares boundaries with adjacent cells (see Fig. \ref{fig:pcd} and  \ref{fig:supergraph}). See \cite[pp. 269-270]{Lav06} for a similar method using vertical lines instead of rays. This decomposition captures which cells occlude others from the survey point $\spt_j$.

\vspace{-10pt}
\subsubsection{Survey-Cell Graph}

\begin{wrapfigure}{r}{0.35\textwidth}
    \centering
    \vspace{-25pt}
    \begin{minipage}{0.35\textwidth}
    \begin{minipage}{0.46\textwidth}
        \centering
        \includegraphics[width=\textwidth]{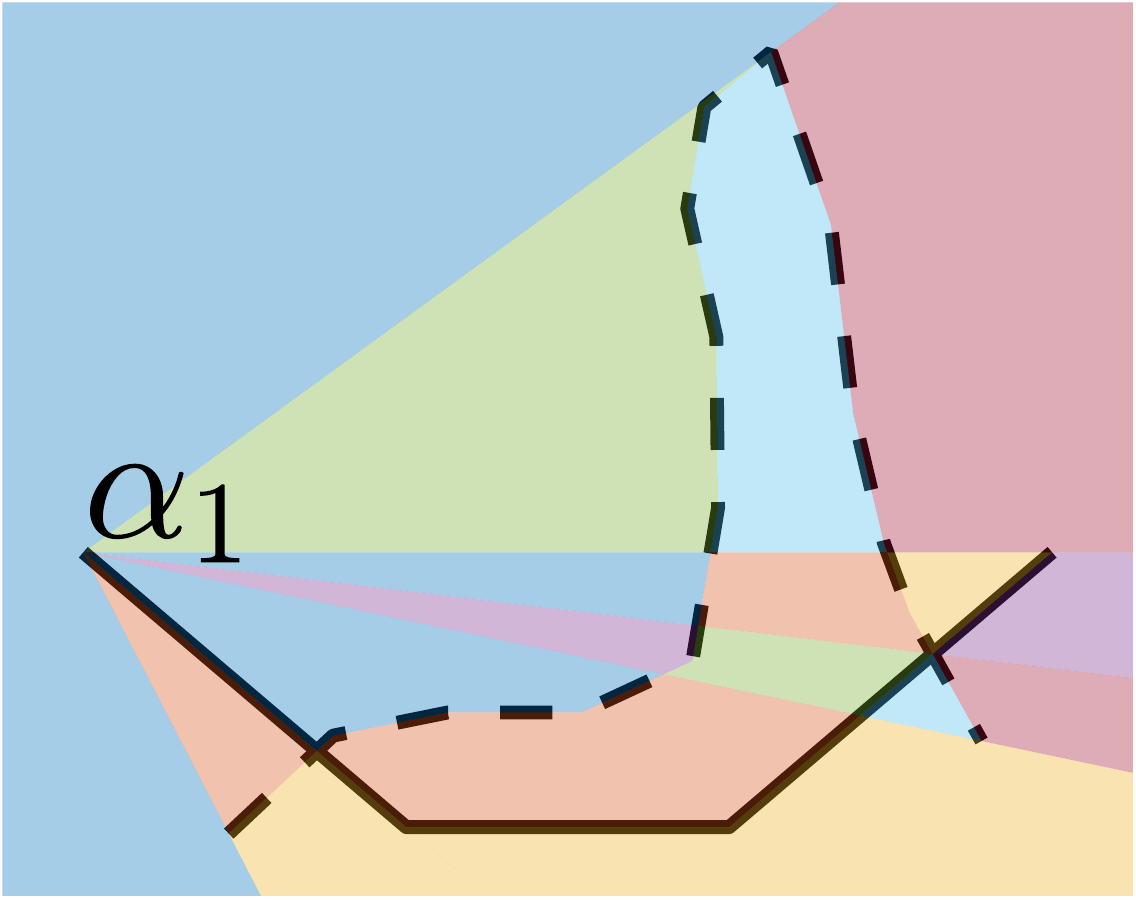}
    \end{minipage}%
    \hfill
    \begin{minipage}{0.46\textwidth}
        \centering 
        \includegraphics[width=\textwidth]{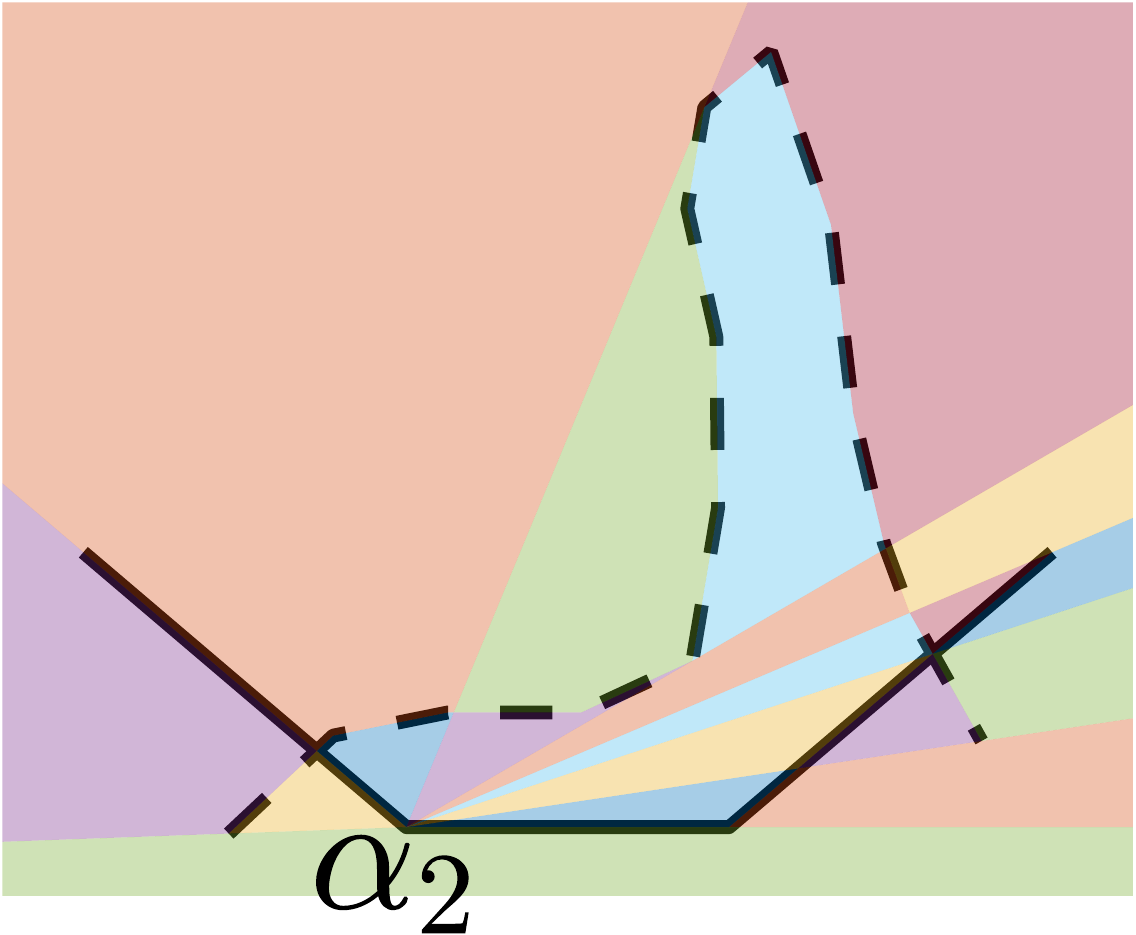}
    \end{minipage}
    \end{minipage}
    \vskip\baselineskip
    \vspace{-9pt}
    \begin{minipage}{0.35\textwidth}
    %\begin{subfigure}[b]{0.32\textwidth}   
        \centering 
        \includegraphics[width=\textwidth]{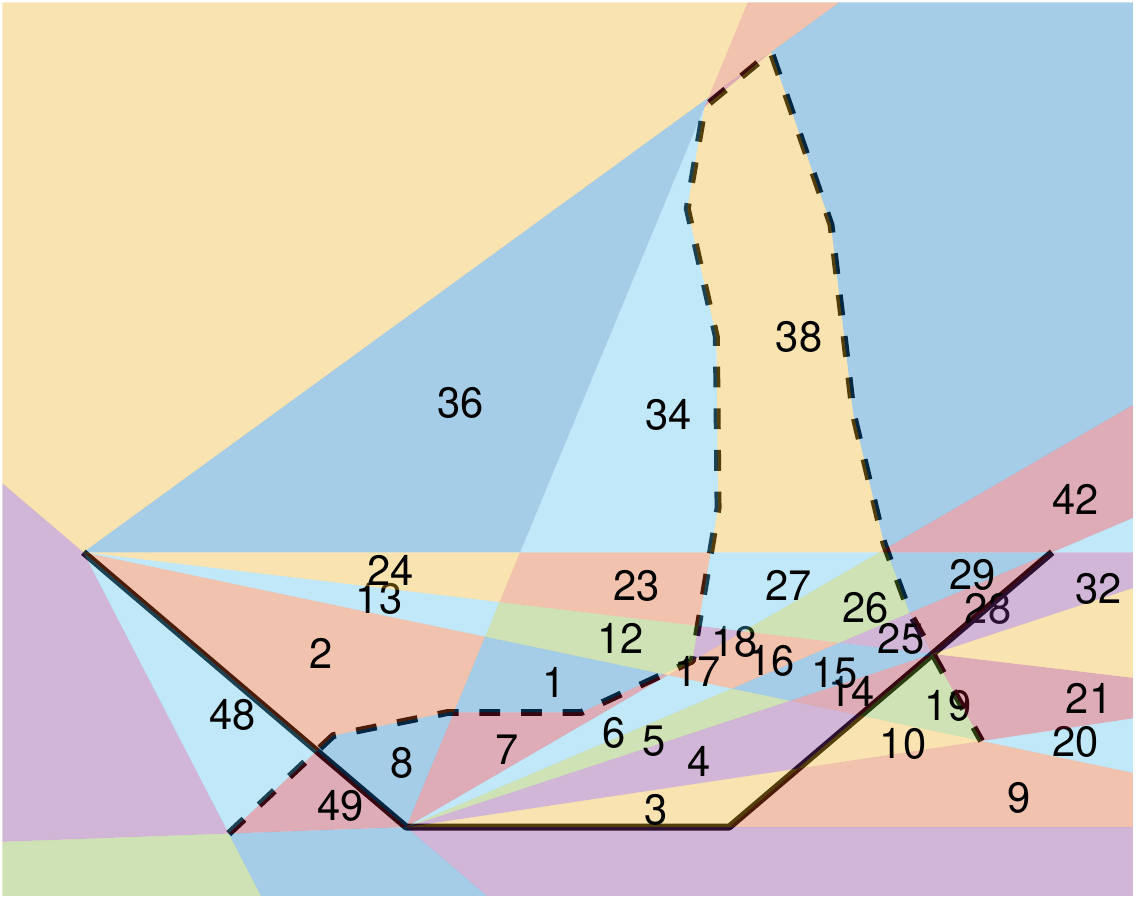}
    %\end{subfigure}
    \end{minipage}\vspace{-5pt}
    \caption{\footnotesize{The full cell decomposition $\fullCell$ composed of the intersection of the two polar cell decompositions $\Cell_{\spt_1}$ and $\Cell_{\spt_2}$. The two individual cell decompositions are shown above.
    }} 
    \label{fig:supergraph}
    \vspace{0pt}
\end{wrapfigure}

To construct the survey-cell graph $G \doteq (N,E)$, we first intersect each cell from each $\Cell_{\spt_j}$ to form the full cell decomposition $\fullCell$ (Fig. \ref{fig:supergraph}). A node $\node = (\spt,\cell) \in N$ of the graph identifies a survey point $\spt \in \SPT$ and a cell $\cell \in \fullCell$. Intuitively, a node corresponds to viewing part of the obstacle in cell $\cell$ from the survey point $\spt$. We use the superscript $\nodesp\node$ to denote the survey point and $\nodecell\node$ to denote the cell and similarly for node sequences and paths. An edge $e \in E$ exists between two different nodes $(\spt,\cell)$ and $(\spt',\cell')$ if the intersection $\mathcal{B} \doteq (\cell \cap \cell') \setminus \safeset$ (recall $\safeset$ is the safe set) is not a single point nor the empty set. Note that any obstacle that lies in $R(t_n)$ corresponds to a path in the survey-cell graph, also denoted a \textit{survey-cell path}%
%(recall $R(t_n)$ is the known region at the third to last vertex of the demonstration)
.

\vspace{-5pt}
\subsubsection{Cell Sequence Identification}\label{sec:class_id}
Prior to formulating the CSP, we identify sets of cell sequences that may contain a solution $O$ to Prob. \ref{prob:cand_traj}. We accomplish this by identifying an upper bound on the cost of the shortest alternative robot path $\xi^i_a$ and a lower bound on the cost of the inferred robot path $\plan\xi^i_{\textrm{inf}}$ from each vertex $x_i$, $i \in \{1, \ldots, n\}$. See App. \ref{app:csqid} for an example.

%\vspace{-5pt}
%\begin{definition}[Cell Sequence Projection]
%Suppose $\class$ is a sequence of cells. Then we define a \textit{cell sequence projection} as $\proj\class \doteq (\bigcup_{\cell \in \class}\cell) \setminus \safeset$.
%\end{definition}
%\vspace{-5pt}

For a sequence of cells $\class$, we will overload $\class$ to also mean the union of cells in $\class$. Now consider a tuple of cell sequences $[\class_2, \ldots, \class_{n+1}]$, and let $\mathbf{O} = \bigcup_i \proj{\class_i}$. Then we find $\bar{c}^i_a \doteq \max_{O \subseteq \mathbf{O}} c(\xi^i_a)$ by planning an alternative (c.f. Sec. \ref{sec:verifying_conf}) with the known obstacle set $\mathbf{O}$. Intuitively, an obstacle that completely fills the cells in $[\class_2, \ldots, \class_{n+1}]$ induces robot paths as long or longer than an obstacle that only fills part of the cells, so $\bar{c}^i_a$ is an upper bound on the cost of the shortest alternative. Furthermore, a lower bound on the cost of the inferred robot path is the length of the robot path from $x_i$ to $x_{i+1}$ to $x_G$ (since the inferred path must always visit the next vertex), denoted $\underline{c}^i_{\textrm{inf}}$. Define $\classid$ such that for all tuples $[\class_2, \ldots, \class_{n+1}] \in \classid$, $\bar{c}^i_a \geq \underline{c}^i_{\textrm{inf}}$ for all $i \in \{1, \ldots, n\}$ (see Fig.~\ref{fig:supergraph}). 
%If a tuple of cell sequences $[\class_2, \ldots, \class_{n+1}] \notin \classid$, then there does not exist an environment configuration $O \subseteq \mathbf{O}$ that solves Prob. \ref{prob:cand_traj} since the alternative robot path will never be more costly than the inferred robot path. 
If $\classid = \emptyset$, then our algorithm guarantees the candidate robot path is unsafe.
    
\vspace{-5pt}
\subsubsection{CSP}\label{sec:csp_comp}

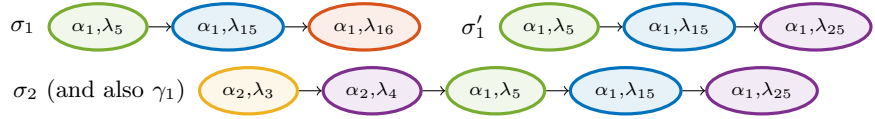
\begin{figure}[t]
    \centering
    \vspace{0pt}
    \begin{tikzpicture}[
    roundnode/.style={ellipse, very thick},
    node distance = 0.3cm
    ]
    %Nodes
    \node[roundnode, draw=matlabGreen,
    fill=matlabGreen!10] (p1_1) {\scriptsize $\spt_1$,$\cell_5$};
    \node[roundnode, draw=matlabBlue, fill=matlabBlue!10] (p1_2) [right=of p1_1]{\scriptsize $\spt_1$,$\cell_{15}$};
    \node[roundnode, draw=matlabOrange, fill=matlabOrange!10] (p1_3) [right=of p1_2]{\scriptsize $\spt_1$,$\cell_{16}$};
    \node (opt1) at (-1,0) {$\graphpath_1$};
    
    \node[roundnode, draw=matlabGreen,
    fill=matlabGreen!10] (p3_1) at (6,0){\scriptsize $\spt_1$,$\cell_5$};
    \node[roundnode, draw=matlabBlue, fill=matlabBlue!10] (p3_2) [right=of p3_1]{\scriptsize $\spt_1$,$\cell_{15}$};
    \node[roundnode, draw=matlabPurple, fill=matlabPurple!10] (p3_3) [right=of p3_2]{\scriptsize $\spt_1$,$\cell_{25}$};
    \node (opt2) at (5,0) {$\graphpath_1'$};
    
    \node[roundnode, draw=matlabYellow, fill=matlabYellow!10] (p2_1) at (2,-0.85) {\scriptsize $\spt_2$,$\cell_3$};
    \node[roundnode, draw=matlabPurple, fill=matlabPurple!10] (p2_2) [right=of p2_1]{\scriptsize $\spt_2$,$\cell_4$};
    \node[roundnode, draw=matlabGreen,
    fill=matlabGreen!10] (p2_3) [right=of p2_2]{\scriptsize $\spt_1$,$\cell_5$};
    \node[roundnode, draw=matlabBlue, fill=matlabBlue!10] (p2_4) [right=of p2_3]{\scriptsize $\spt_1$,$\cell_{15}$};
    \node[roundnode, draw=matlabPurple, fill=matlabPurple!10] (p2_5) [right=of p2_4]{\scriptsize $\spt_1$,$\cell_{25}$};
    \node (obs2) [left=0.05cm of p2_1]{$\graphpath_2$ (and also $\graphpathvar_1$)};
     
    %Lines
    \draw[->] (p1_1.east) -- (p1_2.west);
    \draw[->] (p1_2.east) -- (p1_3.west);
    \draw[->] (p3_1.east) -- (p3_2.west);
    \draw[->] (p3_2.east) -- (p3_3.west);
    
    \draw[->] (p2_1.east) -- (p2_2.west);
    \draw[->] (p2_2.east) -- (p2_3.west);
    \draw[->] (p2_3.east) -- (p2_4.west);
    \draw[->] (p2_4.east) -- (p2_5.west);
    %\draw[style=help lines] (-2,1) grid (8,-2);
    \end{tikzpicture}
    \vspace{-10pt}
    \caption{Three survey-cell graphs representing two obstacles. The colors of the nodes here match the colors of the cells in Fig. \ref{fig:supergraph}. Combining $\graphpath_1$ and $\graphpath_2$ would not produce a valid path as it would branch from node $(\spt_1, \cell_{15})$ to both $(\spt_1$,$\cell_{16})$ and $(\spt_1$,$\cell_{25})$. However, $\graphpath_1'$ and $\graphpath_2$ do chain to form a single path $\graphpathvar_1$ which is the same as $\graphpath_2$. %However, in obstacle recovery, there is an additional point constraint that the combined obstacle must visit both the second and third vertex of the demonstration.
    }
    \label{fig:chaining}
\end{figure}

On the survey-cell graph $G$, we seek to find an \textit{assignment} $\assign$ of $n$ survey-cell paths $(\graphpath_2, ..., \graphpath_{n+1})$ such that we can construct $n$ corresponding obstacles. By Asm. \ref{assume:n_curves}, there can be at most $n$ obstacles in the environment. Fewer than $n$ obstacles manifest in the survey-cell graph as $n$ paths with some that overlap. However, we place some requirements on the paths that overlap in order to ensure that the resulting obstacles do not occlude one another and do not intersect. This is done by virtue of \textit{chaining}; see the definitions below and Fig.~\ref{fig:chaining} for examples of survey-cell paths that can be chained or not.

\vspace{-5pt}
\begin{definition}[Chainable]
Two survey-cell paths $\graphpath$ and $\graphpath'$ are \textit{chainable} if there exists a survey-cell path $\graphpathvar$ such that there is exactly one subsequence of nodes that matches the node sequences of $\graphpath$ and $\graphpath'$ (or their reverse) and each node of $\graphpathvar$ belongs to at least one of these subsequences. Let $\chainable{\graphpath}{\graphpath'}$ be a proposition that is true in this case and otherwise false.
\end{definition}
\vspace{-5pt}

\vspace{-5pt}
\begin{definition}[Obstacle Curve Set]
We denote $\classobs\class$ as the set of obstacle curves belonging to a sequence of cells $\class$, i.e. $o \in \classobs\class$ is a curve that satisfies $o \cap \cell \neq \emptyset$ for all $\cell \in \class$ and $o \cap (\cell \setminus \class) = \emptyset$ for all $\cell \notin \class$.
\end{definition}
\vspace{-5pt}

Consider two survey-cell paths $\graphpath$ and $\graphpath'$ on $G$ that visit the same cell.
%If two survey-cell paths share an overlapping sequence,
%, i.e. they lie in the same class seen from the same survey points, % space saver
Then one of two cases can occur: (i) there may exist two corresponding obstacle curves $o \in \classobs{\graphpath^\cell}, o' \in \classobs{\graphpath'^\cell}$ such that neither obstacle occludes the another from any corresponding survey point. For instance, if two survey-cell paths terminate at the same node $\node$, then the two obstacles curves could both lie in the same cell with a gap between them (from the perspective of the survey point) %, i.e. there exists some ray $r$ from $\nodesp\node$ such that $r \cap \nodecell\node \neq \emptyset$ but $r \cap o_1 = \emptyset$ and $r \cap o_2 = \emptyset$
or (ii) there may not exist any obstacle curves $o$, $o'$ such that one does not occlude the other. In this case, we constrain the resulting obstacle curves to be combined into one obstacle curve so that both can be seen from the same survey points. If they must be combined, we enforce that the paths $\graphpath$ and $\graphpath'$ are chainable.

As each obstacle begins at an intermediate vertex of the demonstration, we require that each path begins at a cell that the vertex lies on its boundary. Formally, let $\startCells{i}$ be the set of cells such that for each $\cell \in \startCells{i}$, $\cell \in \fullCell$ and $x_i \in \partial\cell$. Let $\graphpath^\cell(1)$ be the cell of the first node on the survey-cell path. Then the CSP is formulated below:

\vspace{-5pt}
\begin{equation*}
\text{Find} 
\quad \graphpath_2, \ldots, \graphpath_{n+1} \quad
\text{subject to}
\end{equation*}
\vspace{-20pt}
\begin{align}
&\forall i, \quad \nodecell\graphpath_i(1) \in \startCells{i} \label{eq:constraint_start} \\
&[\nodecell\graphpath_2, \ldots, \nodecell\graphpath_{n+1}] \in \classid \label{eq:constraint_class_id} \\
&\begin{aligned}
    &\forall i, i' \enspace \nexists \node \in \graphpath_{i'} \suchthat \forall o \in \classobs{\nodecell\graphpath_i}, \enspace \fullyocc{\nodecell\node}{o}{\nodesp\node}
\end{aligned} \label{eq:constraint_vision_complete}\\
%&\begin{aligned}
%    &\forall i,j \enspace \text{if} \enspace \exists \node \in \graphpath_{i}, \graphpath_j \suchthat \forall o \in \classobs{\nodecell{\graphpath_i}} \enspace \fullyocc{\nodecell\node}{o}{\nodesp\node} \enspace \text{then} \enspace \chainable{\graphpath_i}{\graphpath_j}
%\end{aligned} \label{eq:constraint_overlapping_complete}\\
&\begin{aligned}
    &\forall i,i' \enspace \text{if} \enspace \exists \node \in \graphpath_i, \graphpath_{i'} \enspace \text{then} \enspace \exists o_i \in \classobs{\graphpath_i^\cell}, o_{i'} \in \classobs{\graphpath_{i'}^\cell} \suchthat \\
    &\qquad (o_i \cap o_{i'} = \emptyset \wedge \neg \occ{o_{i'}}{o_i}{\node^\spt} \wedge \neg \occ{o_i}{o_{i'}}{\node^\spt}) \vee \chainable{\graphpath_i}{\graphpath_{i'}}
\end{aligned}\label{eq:constraint_overlapping_complete}
\end{align}

Constraint \eqref{eq:constraint_start} requires that each survey-cell path starts at a cell containing the vertex $x_{i}$. Constraint \eqref{eq:constraint_class_id} enforces that the resulting survey-cell paths lie in $\classid$, where $\classid$ is the set where there exists a higher cost alternative (c.f. Sec. \ref{sec:class_id}). Constraint \eqref{eq:constraint_vision_complete} enforces that no obstacle curve will always be occluded by another obstacle curve from the respective survey point. This is needed in order to ensure that obstacles are revealed at appropriate points in the demonstration. As discussed earlier, constraint \eqref{eq:constraint_overlapping_complete} enforces that two survey-cell paths that share the same node either chain or contain two obstacles that do not occlude one another in order to ensure that obstacles can be connected if needed. We note \eqref{eq:constraint_vision_complete} and \eqref{eq:constraint_overlapping_complete} can be checked without sampling obstacle curves $o$ since we know what cell the endpoints of obstacles must lie in and therefore can calculate the occluded space for that particular cell sequence. 

The CSP can return failure in finite time in the case that constraint \eqref{eq:constraint_class_id} prevents any otherwise possible assignment. In this case, we return that the candidate robot path is unsafe. However, the CSP may also admit infinite length paths, such as cycling between two nodes with different survey points but the same cell, which prevents the CSP from terminating. We address this shortcoming in the tractable version of the CSP in Sec. \ref{sec:method2}.

\vspace{-10pt}
\subsubsection{Obstacle Recovery}\label{sec:recovery_complete}

\begin{figure}
    \centering
    \vspace{-5pt}
    \begin{minipage}{0.65\textwidth}
    \scalebox{1}{
    \begin{tikzpicture}[
    roundnode/.style={ellipse, very thick},
    node distance = 0.3cm
    ]
    %Nodes
    \node[roundnode, draw=matlabYellow, fill=matlabYellow!10] (p1) {\scriptsize $\spt_2$,$\cell_3$};
    \node[roundnode, draw=matlabPurple, fill=matlabPurple!10] (p2) [below=of p1]{\scriptsize $\spt_2$,$\cell_4$};
    \node[roundnode, draw=matlabGreen,
    fill=matlabGreen!10] (p3) [below=of p2]{\scriptsize $\spt_1$,$\cell_5$};
    \node[roundnode, draw=matlabSkyBlue, fill=matlabSkyBlue!10] (p4) [below=of p3]{\scriptsize $\spt_1$, $\cell_6$};
    \node[roundnode, draw=matlabRed, fill=matlabRed!10] (p5) [right=of p4] {\scriptsize $\spt_1$, $\cell_7$};
    \node[roundnode, draw=matlabYellow, fill=matlabYellow!10] (p6) [above=of p5]{\scriptsize $\spt_1$, $\cell_{17}$};
    \node[roundnode, draw=matlabOrange, fill=matlabOrange!10] (p7) [above=of p6]{\scriptsize $\spt_1$, $\cell_{16}$};
    \node[roundnode, draw=matlabBlue, fill=matlabBlue!10] (p8) [above=of p7]{\scriptsize $\spt_1$,$\cell_{15}$};
    \node[roundnode, draw=matlabPurple, fill=matlabPurple!10] (p9) [right=of p8]{\scriptsize $\spt_1$,$\cell_{25}$};
    \node[roundnode, draw=matlabGreen, fill=matlabGreen!10]
    (p10) [below=of p9]{\scriptsize $\spt_1$,$\cell_{26}$};
    \node[roundnode, draw=matlabSkyBlue, fill=matlabSkyBlue!10]
    (p11) [below=of p10]{\scriptsize $\spt_1$,$\cell_{27}$};
    \node[roundnode, draw=matlabYellow, fill=matlabYellow!10] (p12) [below=of p11]{\scriptsize $\spt_1$,$\cell_{38}$};
     
    %Lines
    \draw[->] (p1.south) -- (p2.north);
    \draw[->] (p2.south) -- (p3.north);
    \draw[->] (p3.south) -- (p4.north);
    \draw[->] (p4.east) -- (p5.west);
    \draw[->] (p5.north) -- (p6.south);
    \draw[->] (p6.north) -- (p7.south);
    \draw[->] (p7.north) -- (p8.south);
    \draw[->] (p8.east) -- (p9.west);
    \draw[->] (p9.south) -- (p10.north);
    \draw[->] (p10.south) -- (p11.north);
    \draw[->] (p11.south) -- (p12.north);
    
    \draw[dashed] (-0.8, 0.4) -- (0.75, 0.4) -- (0.75, -1.35) -- (-0.8, -1.35) -- (-0.8, 0.4);
    \draw[dashed] (-0.8, -1.45) -- (-0.8, -3.3) -- (4.4, -3.3) -- (4.4, 0.4) -- (0.9, 0.4) -- (0.9,-1.45) -- (-0.8, -1.45);
    
    \draw [decorate,decoration={brace,amplitude=10pt,mirror},xshift=-2pt,yshift=0pt] (-0.8,0.4) -- (-0.8,-1.35) node [black,midway,xshift=-0.7cm] {$\seg_{1,1}$};
    \draw [decorate,decoration={brace,amplitude=10pt},xshift=2pt,yshift=0pt] (4.4,0.4) -- (4.4,-3.3) node [black,midway,xshift=0.7cm] {$\seg_{1,2}$};
    
    %\draw[help lines] (-1,1) grid (5,-4);
    \end{tikzpicture}
    }
    \centering
    Survey-cell path $\graphpathvar_1$
    \end{minipage}%
    %\hspace{5pt}
    \begin{minipage}{0.35\textwidth}
    \includegraphics[width=\textwidth]{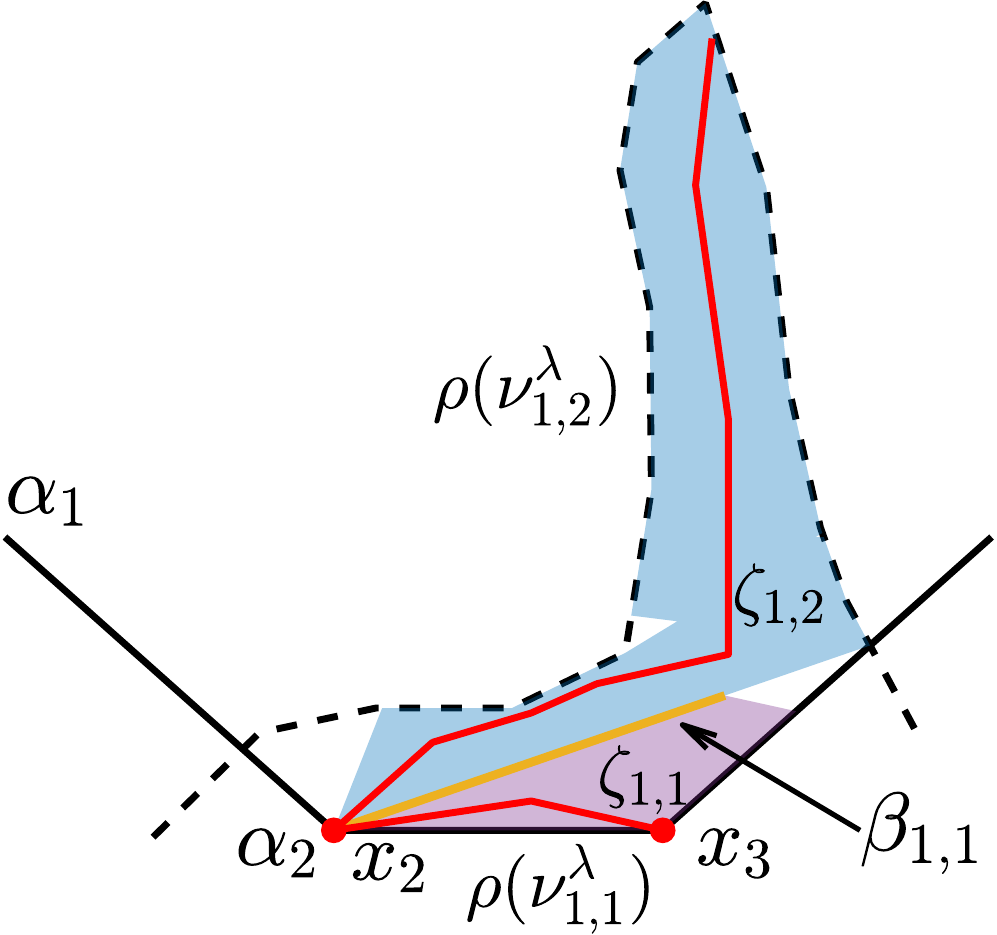}
    \centering
    \end{minipage}
    \vspace{-12.5pt}
    \caption{On the left is shown a survey-cell path $\graphpathvar_1$ on the survey-cell graph corresponding to Fig. \ref{fig:supergraph}. The path is partitioned into two survey-cell segments, $\seg_{1,1}$ and $\seg_{1,2}$, each with a unique survey point. On the right is shown the candidate (dashed black), demonstration (solid black), and cell sequences corresponding to the survey-cell path segments in blue and purple. The corresponding obstacle segments $\obsseg_{1,1}$ and $\obsseg_{1,2}$ (in red) must intersect at a join point on the yellow line. This obstacle is also required to intersect with two points constraints in red, $x_2$ and $x_3$.}
    \label{fig:representation}
    \vspace{12.5pt}
\end{figure}
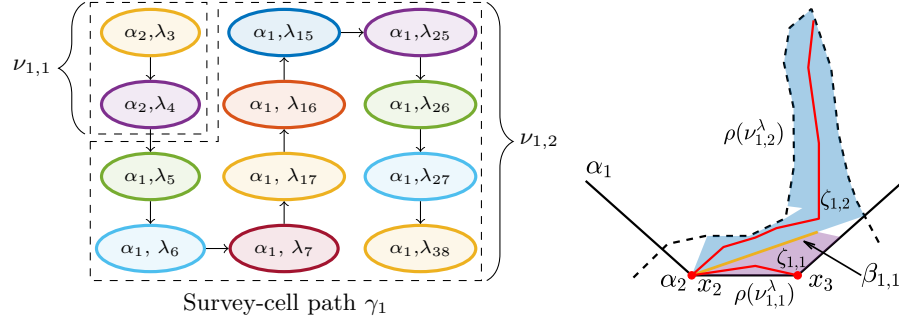

For sake of brevity, we describe obstacle recovery at a high level here (see a detailed explanation in App. \ref{app:obs_rec}). We seek to find $m \leq n$ obstacles in $\X$ (curves parameterized by line segments) from the assignment $\assign$. We will first chain survey-cell paths if necessary, producing a reduced set of survey-cell paths. For the reduced set $\{\graphpathvar_1, \ldots, \graphpathvar_m\}$, we partition each survey-cell path $\graphpathvar_\kappa$ into the minimum number of nonoverlapping survey-cell paths $\seg_{\kappa,1}, \ldots, \seg_{\kappa,K_\kappa}$ such that each path is only seen from one survey point (see Fig.~\ref{fig:representation}). 

Corresponding to each of these paths $\seg_{\kappa,l}$, we sample an obstacle segment $\obsseg_{\kappa,l}$ by sampling lines that are entirely visible to the survey point $\nodesp{\seg_{\kappa,l}}$ and lie in the cell sequence $\proj{\seg_{\kappa,l}}$. 
%In general, any sampling method that generates a radial function with origin $\seg_{\kappa,l}^\spt$ and lies in $\seg_{\kappa,l}$ will produce a segment entirely visible to the survey point. 
This produces $m$ obstacles such that $o_\kappa = \bigcup_{l} \obsseg_{\kappa,l}$ for $\kappa \in \{1, \ldots, m\}$. To handle sampling obstacles with an arbitrary number of line segments, we require our obstacle segment sampler to be capable of sampling curves with a specified number of line segments. Define a \textit{segment discretization} as $\segdisc \doteq \{k_{1,1}, \ldots, k_{m,K_m}\}$. To sample an obstacle configuration $O$, we sample each obstacle segment $\obsseg_{\kappa,l}$ such that $\obsseg_{\kappa,l}$ consists of $k_{\kappa,l}$ line segments and satisfies two constraints: the resulting obstacle curves $o_\kappa$ are continuous and visit the necessary vertices of the demonstration (also called \textit{point constraints}). In brief, we enforce continuity by restricting adjacent obstacle segments $\obsseg_{\kappa,l}$ and $\obsseg_{\kappa,l+1}$ to intersect at a \textit{join point} $\joinpt_{\kappa,l}$.

Probabilistically-Complete Obstacle Sampling is summarized in Alg. \ref{alg:m1_find_environment_conf}. The function \texttt{AssignShorterThan}$(K_1)$ finds all assignments of the CSP with survey-cell paths no longer than $K_1$. \texttt{DiscLessThan}$(K_2)$ gives all discretizations $\segdisc$ such that $k \leq K_2$ for all $k \in \segdisc$. \texttt{SampleObs} follows the procedure described above and \texttt{Valid}$(\env)$ tests if $\env$ is valid according to Def. \ref{def:consistency}.

%\begin{wrapfigure}{l}{1\textwidth}
%\newlength{\textfloatsepsave} 
%\setlength{\textfloatsepsave}{\textfloatsep}
\setlength{\textfloatsep}{5pt}
\begin{algorithm}[t]
\DontPrintSemicolon
\SetAlgoLined
\LinesNumbered

\SetKwFunction{AssignmentsShorterThan}{AssignShorterThan}
\SetKwFunction{DiscLessThan}{DiscLessThan}
\SetKwFunction{SampleObs}{SampleObs}
\SetKwFunction{Valid}{Valid}
\SetKwFunction{FullCell}{FullCellDecomposition}
\SetKwFunction{ConstructGraph}{ConstructGraph}
\SetKwFunction{ClassID}{CellSeqID}
\SetKwFunction{ConstructCSP}{ConstructProbCompCSP}

\KwIn{$\demo$, $\candtraj$, $\SPT$} \KwOut{possibly safe or unsafe}
$\fullCell \leftarrow $ \FullCell{$\demo$, $\candtraj$, $\SPT$} \\
$G \leftarrow $ \ConstructGraph{$\fullCell$, $\SPT$}, 
$\classid \leftarrow $ \ClassID{$G$, $\demo$} \\
\lIf{$\classid = \emptyset$}{%
    \Return{unsafe}%
}
\texttt{CSP} $\leftarrow$ \ConstructCSP{$G$, $\classid$}, $K_1 \leftarrow 1$, $K_2 \leftarrow 1$ \\
\While{True}{
    $\assign_1, .., \assign_{N_1} \leftarrow$ \texttt{CSP.}\AssignmentsShorterThan{$K_1$} \\
    \For{$n_1 \leftarrow 1$ \KwTo $N_1$}{
        $\segdisc_1, .., \segdisc_{N_2} \leftarrow $ \DiscLessThan{$K_2$} \\
        \For{$n_2 \leftarrow 1$ \KwTo $N_2$}{
            $O \leftarrow $ \SampleObs{$\assign_p$, $\segdisc_q$} \\
            \lIf{\Valid{$O$}}{\Return{possibly safe}}
        }
    }
    $K_1 \leftarrow K_1 + 1$, $K_2 \leftarrow K_2 + 1$
}
\caption{ProbComp Obstacle Sampling}
\label{alg:m1_find_environment_conf}
\end{algorithm}
%\end{wrapfigure}

\vspace{-8pt}
\subsection{Analysis of Algorithm \ref{alg:m1_find_environment_conf}}\label{sec:prob_comp}
\vspace{-3pt}

Suppose there exists $\env$ such that $\demo$ solves Prob. \ref{prob:arrive_at_goal} and $\env$ satisfies Asm. \ref{assume:line_segments} - \ref{assume:n_curves}. Suppose we can perturb each vertex of $O$ (other than those coincident with a vertex of the demonstration) within an $\epsilon$-ball that results in a set of consistent configurations $\solnenv$ that also satisfy our assumptions.

\vspace{-5pt}
\begin{theorem}[Probabilistic Completeness]\label{th:prob_comp}
Let $\mathbb{O}_{n_1}$ be the set of samples drawn by the $n_1^{th}$ iteration of the outer loop of Alg. \ref{alg:m1_find_environment_conf}. Then, under Asm. \ref{assume:line_segments} - \ref{assume:n_curves} and for some $0 < K < \infty$ and $0 < \epsilon_2 < 1$,

\vspace{-8pt}
\begin{equation}
\mathbb{P}(\mathbb{O}_{n_1} \cap \solnenv \neq \emptyset) \geq 1 - (1-\epsilon_2)^{n_1 - K} \qquad \forall n_1 \geq K
\end{equation}
\end{theorem}
\vspace{-5pt}

As a consequence to Thm. \ref{th:prob_comp}, $n_1 \to \infty$ implies $\mathbb{P}(\mathbb{O}_{n_1} \cap \solnenv \neq \emptyset) = 1$.

\vspace{-5pt}
\begin{theorem}[Candidate Unsafe]\label{th:cand_unsafe}
Suppose Assumption \ref{assume:line_segments} and \ref{assume:n_curves} is satisfied and $\classid = \emptyset$. Then the candidate robot path $\candtraj$ is unsafe.
\end{theorem}
\vspace{-5pt}

\vspace{-10pt}
\subsection{Heuristic-guided Obstacle Sampling}\label{sec:method2}
\vspace{-5pt}

Alg. \ref{alg:m1_find_environment_conf} is intractable for a number of reasons, including the infinite length paths already mentioned. As another example, the number of discretizations with increasing $K_2$ grows factorially. To make the method tractable, we introduce and modify constraints of the CSP to speed the search for a viable assignment. Furthermore, we present a method for efficiently parameterizing obstacle segments and introduce a gradient ascent procedure. These changes improve tractability but sacrifice completeness of the method.

\vspace{-8pt}
\subsubsection{Tractably Handling Occlusions}

Though constraint \eqref{eq:constraint_vision_complete} captures exactly what survey-cell paths necessarily contain an obstacle that occludes others, calculating occlusions can be survey-cell path dependent and we empirically found it to be more efficient to consider partial occlusions instead of full occlusions. This change necessitates the addition of a \textit{pseudolayer} (defined below). This pseudolayer captures instances of survey-cell paths where an obstacle partially occludes other cells in the path, but there still exists an obstacle that lies in the obstacle curve set that does not self occlude.

\vspace{-5pt}
\begin{definition}[Pseudolayer] Suppose a survey-cell path $\graphpath_i$ (or its reverse) has a subsequence of nodes $\mathcal{N}$ such that the first node $\node_1$ partially occludes the last node $\node_2$ from the last node's survey point $\spt$. This subsequence is a pseudolayer if there exists a ray $r$ originating from the survey point $\spt$ such that $r \cap \node \neq \emptyset$ for all $\node \in \mathcal{N}$. In this case, we say $\adj{\node_1}{\node_2}{i}$ is true and otherwise false.
%\todo{this definition is hard to parse but the example next might  help. will check the example in the next pass.}
\end{definition}
\vspace{-5pt}

For example in Fig.~\ref{fig:supergraph}, the cells 18, 16, and 15 form a pseudolayer with respect to $\spt_1$ but cells 18, 27, 26, 25, and 15 do not. Furthermore, we require a little more notation before defining the tractable CSP. If $\class$ is a sequence, let $\class(k)$ be the $k^{th}$ entry in the sequence. Recalling $\SPT$ is the set of survey points, let $\SPT(x_i)$ be the set of survey points on the demonstration before the point $x_i$. Then,

\vspace{-5pt}
\begin{equation*}
\text{Find} 
\quad \graphpath_2, \ldots, \graphpath_{n+1} \quad
\text{subject to}
\end{equation*}
\vspace{-22.5pt}
\begin{align}
&\text{Constraints \eqref{eq:constraint_start} and \eqref{eq:constraint_class_id}} \nonumber \\
&\begin{aligned}
    &\forall i, {i'}, \node_1 \in \graphpath_i, \nexists \node_2 \in \graphpath_{i'} \suchthat \occ{\nodecell\node_2}{\nodecell\node_1}{\nodesp\node_2} \wedge
    (i \neq {i'} \vee \neg\adj{\node_1}{\node_2}{i})
\end{aligned} \label{eq:constraint_vision}\\
&\forall i,{i'} \enspace \text{if} \enspace \exists \node \in \graphpath_i, \graphpath_{i'} \enspace \text{then} \enspace \chainable{\graphpath_i}{\graphpath_{i'}} \label{eq:constraint_overlapping} \\
&\forall i,j \enspace \spt_j \in \nodesp{\graphpath_i} \Rightarrow \spt_j \in \SPT(x_{i+1}) \label{eq:constraint_future} \\
&\forall i \enspace \nexists k_1, k_2, k_1 \neq k_2 \suchthat \nodecell{\graphpath_i}(k_1) = \nodecell{\graphpath_i}(k_2) \label{eq:no_cycles}\\
&\forall i,{i'}, \node_1 \in \graphpath_i, \nexists \node_2 \in \graphpath_{i'} \suchthat \nodecell{\node_1} =\nodecell{\node_2}, \nodesp{\node_1} \neq \nodesp{\node_2} \label{eq:shared_cells}
\end{align}
\vspace{-10pt}

We modify constraint \eqref{eq:constraint_vision_complete} into \eqref{eq:constraint_vision} so that if any cell in a survey-cell path occludes a particular cell $\cell$ from a survey point $\spt$, then the corresponding node $\node = (\spt, \cell)$ cannot lie on any path unless they lie in a pseudolayer. In this case, it still may be possible to construct an obstacle curve that does not occlude itself. We modify the overlapping constraint from \eqref{eq:constraint_overlapping_complete} into \eqref{eq:constraint_overlapping} so that all overlapping paths are chained. We include \eqref{eq:constraint_future} to ensure no obstacle is seen by a future survey point%,
% which drastically reduces the search space
. We include \eqref{eq:no_cycles} to eliminate paths that visit the same cell multiple times. Finally, $\eqref{eq:shared_cells}$ is included so that no cell is seen from different survey points.
%we construct can lie anywhere in a cell without potentially intersecting other obstacle curves.

\vspace{-8pt}
\subsubsection{Specialized Obstacle Recovery}\label{sec:convex_grad}
\vspace{-3pt}

We briefly describe the specialized obstacle sampling here and leave a more detailed discussion for App. \ref{app:special_obs_rec}. Essentially, we perform a convex decomposition of each cell sequence of a survey-cell segment $\seg_{\kappa,l}$ by intersecting rays from the corresponding survey point $\seg^\spt_{\kappa,l}$. This, along with join points and point constraints, forms a parameterization $\theta$. We sample $\theta$ and optionally perform gradient ascent to arrive at a consistent obstacle.
%We can furthermore improve performance by incorporating a gradient method based on the \textit{locally supporting} vertices of the obstacle (obstacle vertices that intersect with the alternative or inferred plans).

\vspace{-8pt}
\subsubsection{Computational Complexity}
\vspace{-3pt}

We present a detailed derivation of the computational complexity in App. \ref{app:comp_complex}. In brief, supposing there are a total of $N_v$ vertices on the demonstration and candidate combined, the running time is dominated by the CSP which considers $O(n (N_\spt^2) \, \hat \, (N_v^{N_\spt}) )$ paths. A key takeaway is that the number of survey points dominates the computational complexity.

\begin{algorithm}[t]
\DontPrintSemicolon
\SetAlgoLined
\LinesNumbered

\SetKwFunction{CSP}{NextAssign}
\SetKwFunction{ObsParam}{ObsParam}
\SetKwFunction{CalculateGrad}{CalculateGrad}
\SetKwFunction{Valid}{Valid}
\SetKwFunction{FullCell}{FullCellDecomposition}
\SetKwFunction{ConstructGraph}{ConstructGraph}
\SetKwFunction{ClassID}{CellSeqID}
\SetKwFunction{ConstructCSP}{ConstructHeuristicCSP}

\KwIn{$\demo$, $\candtraj$, $\SPT$}
\KwOut{possibly safe, guaranteed unsafe, undecided}

$\fullCell \leftarrow $ \FullCell{$\demo$, $\candtraj$, $\SPT$} \\
$G \leftarrow $ \ConstructGraph{$\fullCell$, $\SPT$}, $\classid \leftarrow $ \ClassID{$G$, $\demo$} \\
\lIf{$\classid = \emptyset$}{%
    \Return{guaranteed unsafe}%
}
\texttt{CSP} $\leftarrow$ \ConstructCSP{$G$, $\classid$}

\For{$i_\textrm{assign} = 1$ \KwTo $\maxassign$}{
    $\assign \leftarrow $ \texttt{CSP}.\CSP{} \\
    \For{$i_\textrm{initial}=1$ \KwTo $\maxinitial$}{
        $\param \leftarrow $ \ObsParam{$\assign$} \\
        \For{$i_\textrm{grad}=1$ \KwTo $\maxgrad$}{
            $\param \leftarrow \param + \grad $ \CalculateGrad{$\param$} \\
            \lIf{\Valid{$O(\param)$}}{\Return{possibly safe}}
        }
    }
}
\Return{undecided}
\caption{Heuristic-Guided Obstacle Sampling}
\label{alg:m2_find_environment_conf}
\end{algorithm}

\vspace{-8pt}
\subsection{Applications in Planning and Scouting}
\vspace{-3pt}
To generate candidate robot paths, we can utilize planners which reason over the safety of entire paths, such as path integral control \cite{Kappen_PI}, cross entropy method \cite{Rubinstein:2004:CEM:1014902}, or hit and run \cite{hit_and_run}. At a high level, each of these methods rely on sampling control actions and iteratively refining to produce candidate robot paths.
%Furthermore, if path refinement is needed after finding an initial candidate, the method can be sped up by checking if the new candidate robot path intersects the previous environment configuration $O$. If it does not, then the new candidate is also safe with respect to $O$. % space saver

To adjust our method for sequential demonstrations $\{\demo_k\}_{k=1}^{K}$, we note that the inferred and alternative plans made from each intermediate vertex of each demonstration must plan to the respective goal of each demonstration. To solve Prob. \ref{prob:env_prob}, instead of terminating our algorithm after finding one feasible environment configuration, we run it for all feasible assignments and sample multiple obstacles for each assignment resulting in environment configurations $\envset$. Different probability measures may be employed based on prior knowledge, or simply uniformly, i.e. $\mathbb{P}(\env) = |\envset|^{-1}$ for all $\env \in \envset$. %Different probability measures may be employed based on prior knowledge, for instance with respect to a particular ``score" of an environment configuration.

\vspace{-8pt}
\section{Experimental Results}
\vspace{-3pt}

We will apply our method to three experiments that focus on planning novel paths in the environment. In the first, we take a candidate, demonstration, and set of survey points and seek to find a consistent environment. This experiment provides examples of obstacles that can be found with our method and situations in which we reject the candidate as certainly unsafe. In the second experiment, we generate a set of candidate trajectories using hit-and-run \cite{hit_and_run} and run our method on each to determine a set of candidates that are possibly safe in some environment. This experiment evaluates the discriminatory ability of our method, i.e. which candidates are identified as guaranteed unsafe, possibly safe, or undecided. We additionally evaluate how many possibly safe candidates are actually safe in the true environment. In the third experiment, we run our method on sequential demonstrations to generate an obstacle ``cloud" or a set of consistent environments. We evaluate the quality of this cloud by comparing its similarity (via an appropriate distance metric) to the true obstacle and by evaluating the safety of novel planned paths.

All experiments were implemented in MATLAB and run on an Intel® Core™ i7-6700 CPU @ 3.40GHz x 8 with 32 GB of RAM. We run Alg. \ref{alg:m2_find_environment_conf} setting $\maxassign = 10$, $\maxgrad = 5$, and $\maxinitial = 4$. In all examples the survey points are all vertices of the demonstration except for the last two.

\vspace{-12pt}
\subsection{Single Candidate}

\begin{figure}
\vspace{-27.5pt}
\begin{minipage}{0.20\textwidth}
    \centering
    \includegraphics[width=\textwidth]{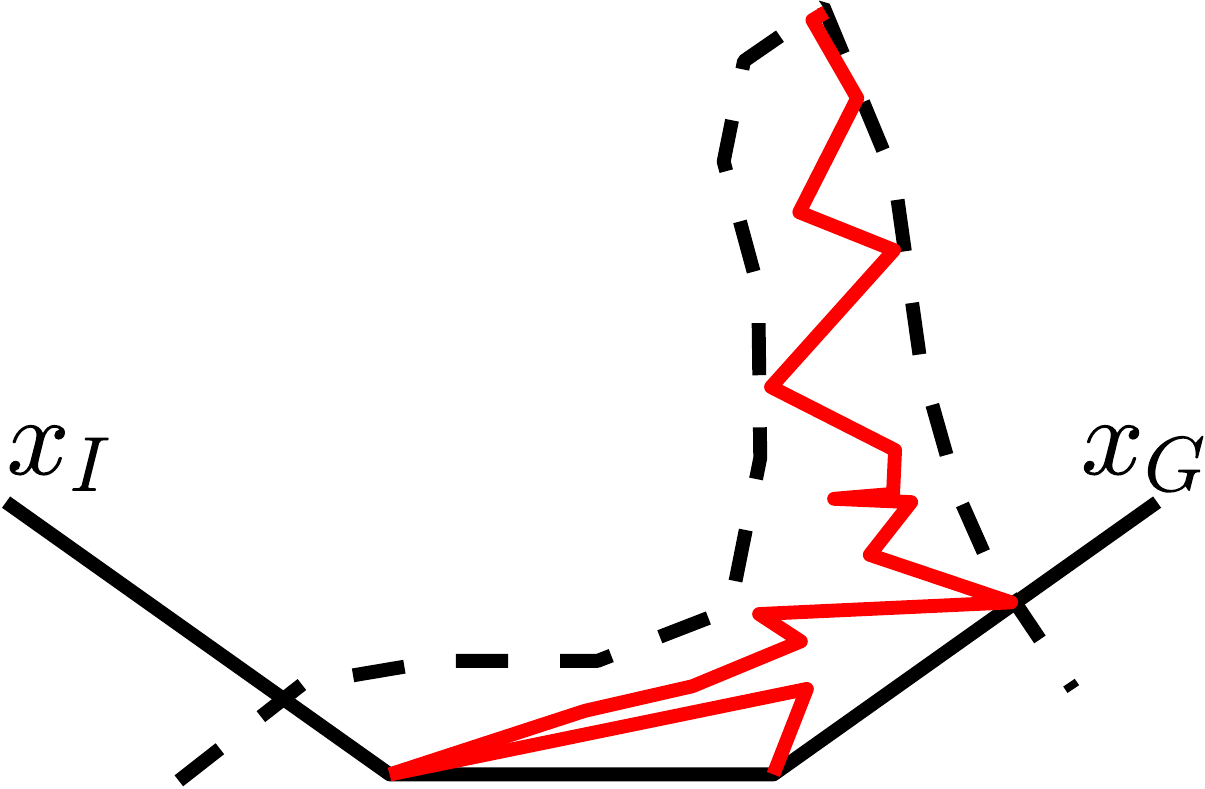}
    (a)
    %\label{fig:obs1}
\end{minipage}%
\hfill
\begin{minipage}{0.20\textwidth}
    \centering
    \includegraphics[width=\textwidth]{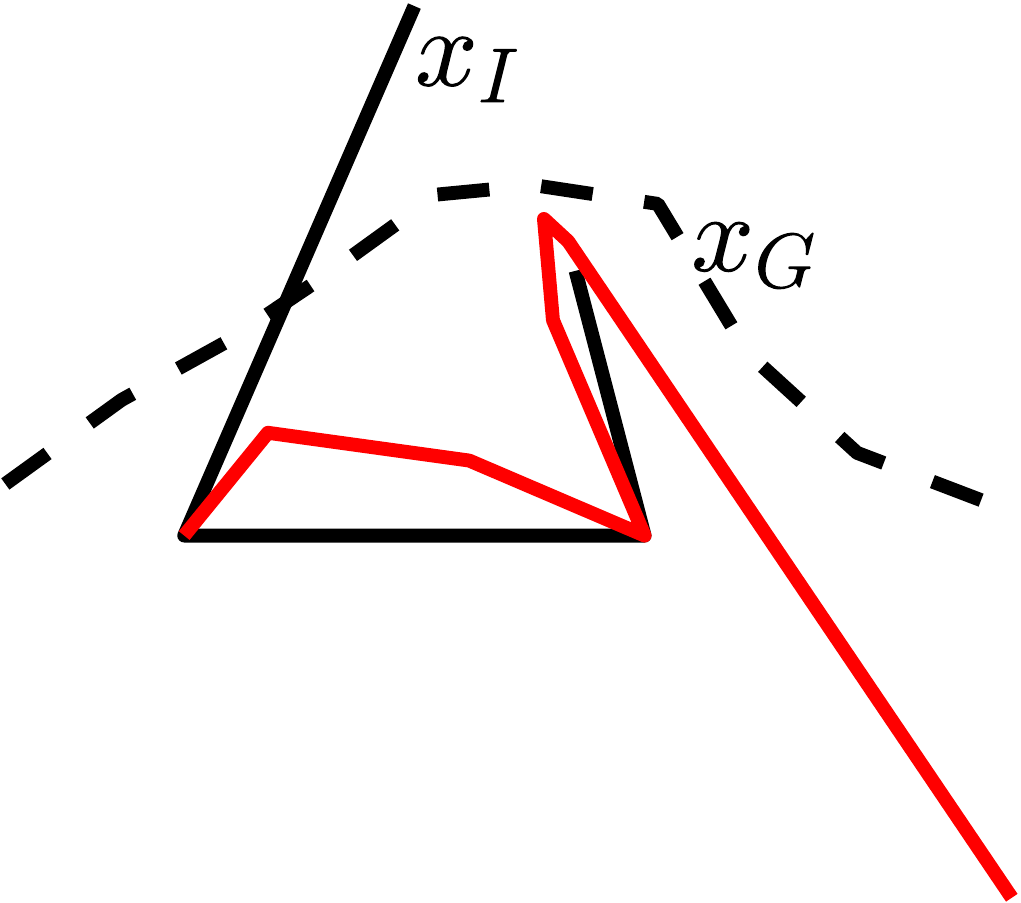}
    (b)
    %\label{fig:obs4}
\end{minipage}%
\hfill
\begin{minipage}{0.20\textwidth}
    \centering 
    \includegraphics[width=\textwidth]{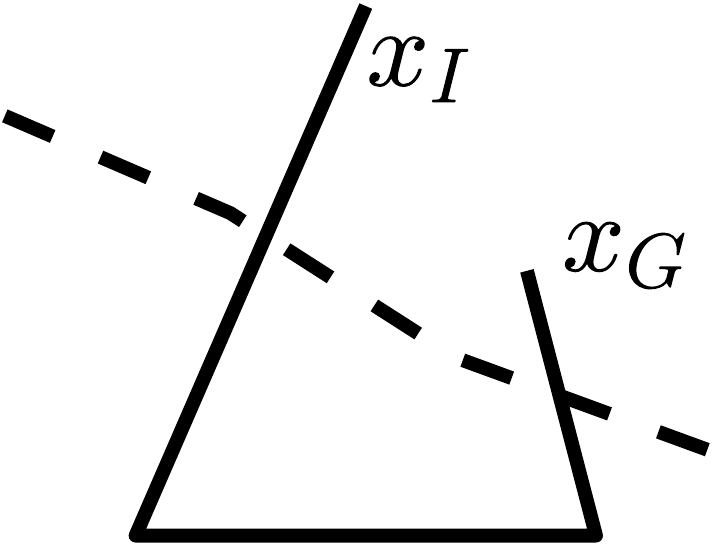}
    (c)
    %\label{fig:unsafe_cand}
\end{minipage}\hfill
\begin{minipage}{0.25\textwidth}
    \centering 
   \begin{tikzpicture}
        \node[anchor=south west,inner sep=0] at (0,0) {\includegraphics[width=\textwidth]{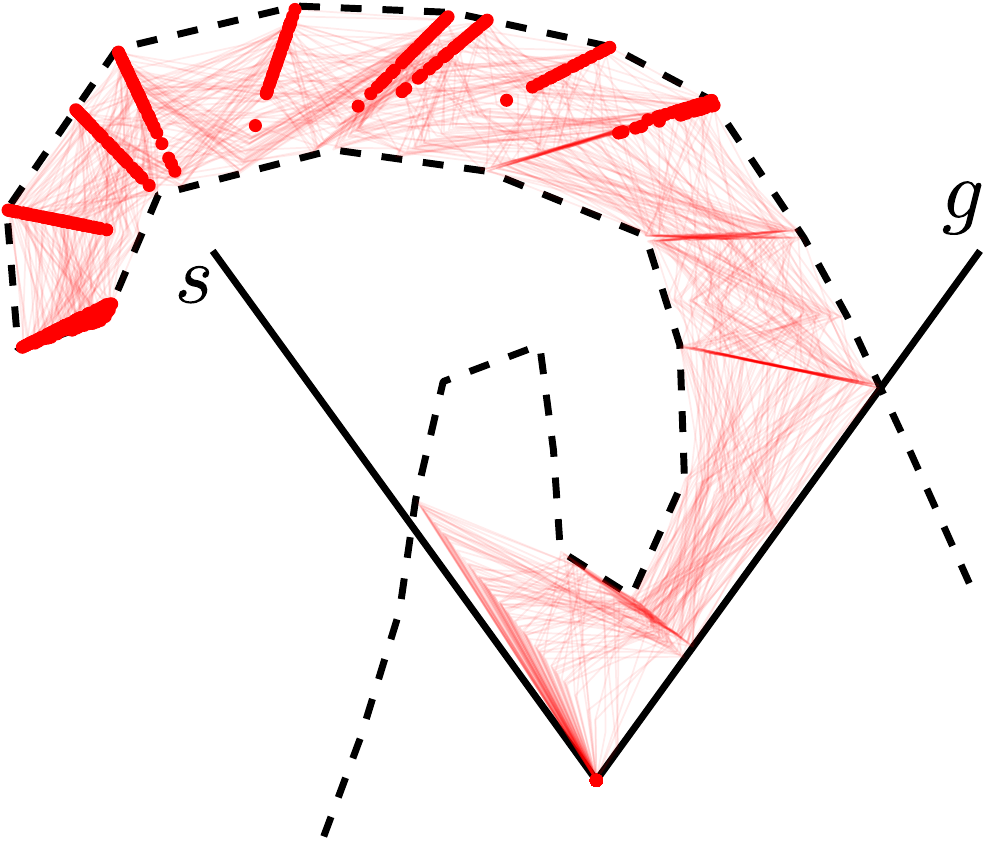}};
        \draw[white,fill=white] (0.6,1.74) circle (0.6ex);
        \node at (0.6,1.62) {$x_I$};
        \draw[white,fill=white] (2.95,2) circle (1ex);
        \node at (2.95,2) {$x_G$};
        %\draw[help lines] (0,0) grid (2,2);
    \end{tikzpicture}
   (d)
    %\label{fig:unsafe_cand}
\end{minipage}
\vspace{-8pt}
    \caption{\small{Example environments with different candidates. Demonstration (solid black), candidate (dashed black), learned obstacle (red). In (d), we show 100 found obstacles with locally supporting vertices shown with a red dot.}} 
    \label{fig:obs_results}
    %\vspace{0pt}
\end{figure}
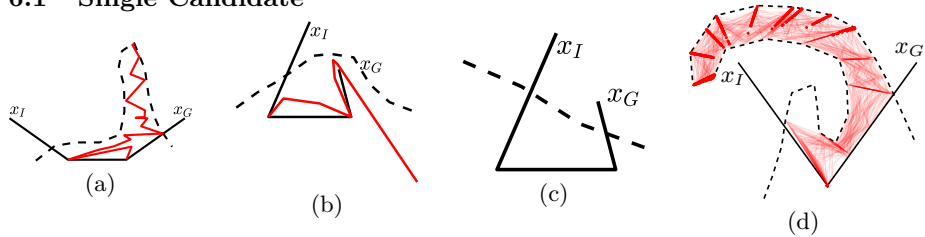

In Fig. \ref{fig:obs_results}, we show demonstrations with different candidates and resulting environment configurations found by our method. We see that it is capable of finding obstacles that coincide with multiple points of the demonstration (Fig. \ref{fig:obs_results}a) rather than finding $n$ separate obstacles. The method can also exploit previously gained knowledge of the demonstrator (Fig. \ref{fig:obs_results}b) since the obstacle is only entirely visible from $x_I$. Furthermore, in Fig. \ref{fig:obs_results}c, our method does not return a configuration by constraint \eqref{eq:constraint_class_id} which is correct by Thm. \ref{th:cand_unsafe}. In Fig. \ref{fig:obs_results}d, we show 100 obstacles found for the given demonstration and candidate with locally supporting vertices for each obstacle shown with a red dot.

\vspace{-10pt}
\subsection{Evaluation of Many Candidates}
 
We consider the problem of planning a novel path to aid an individual inside of a house where access to a map is forbidden for privacy reasons. However, we have access to demonstration provided by a previous medical team. Rather than plan a single path that may be safe, our goal is to plan many possibly safe paths.
%and select a plan online when sensing information is available.

\begin{wrapfigure}{r}{0.35\textwidth}
\vspace{0pt}
    \centering
    \includegraphics[width=0.35\textwidth]{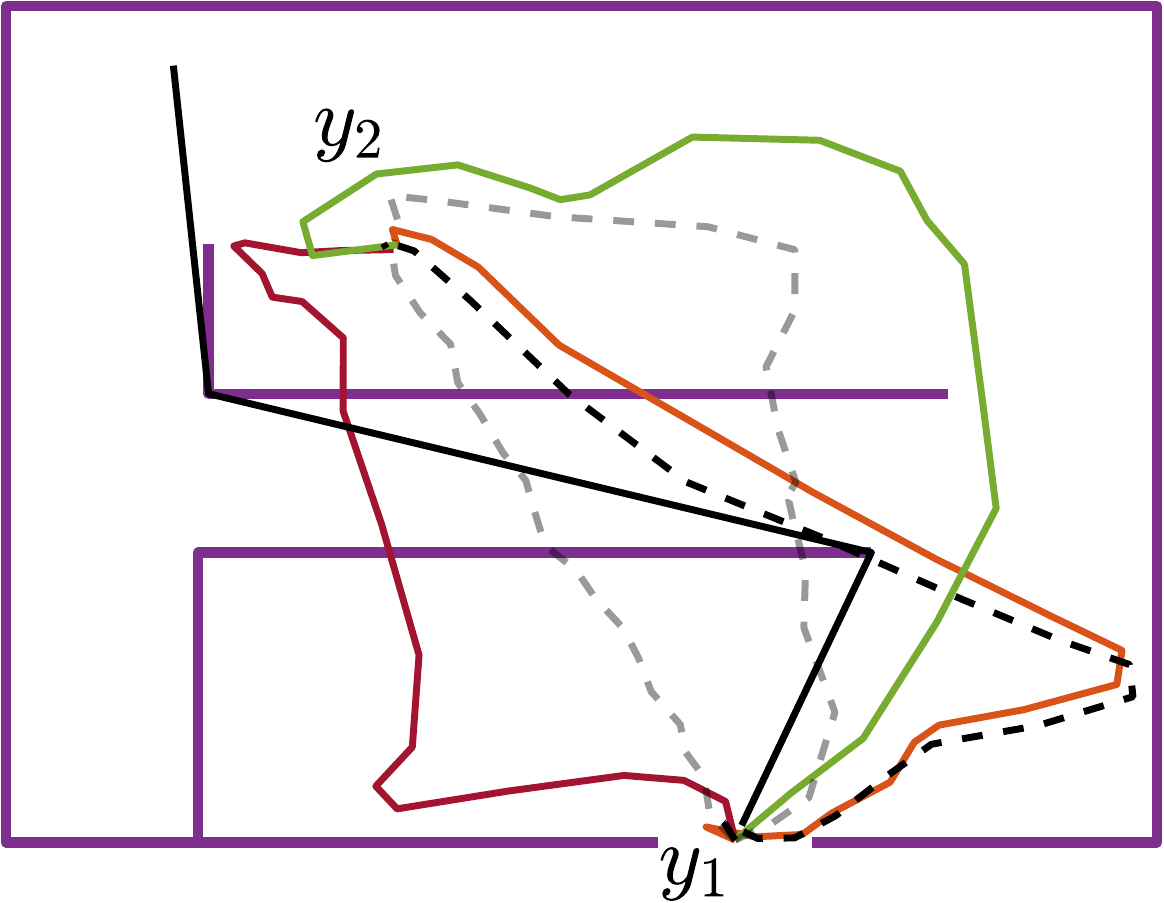}
    %\label{fig:planning}
    \vspace{-26pt}
     \caption{\footnotesize{An example environment of a house (purple), where we seek to plan a path to position $y_2$. Demonstration is solid black. Our method finds three possibly safe candidate paths (red, orange, green). 
     %The green path is safe with respect to the true environment. 
     Also shown are unsafe candidate robot paths (dashed gray and black).}}
     \vspace{-18pt}
     \label{fig:cloud_planning}
\end{wrapfigure}

\begin{wrapfigure}{r}{0.35\textwidth}
    \vspace{-22pt}
    \centering
    \includegraphics[width=0.33\textwidth]{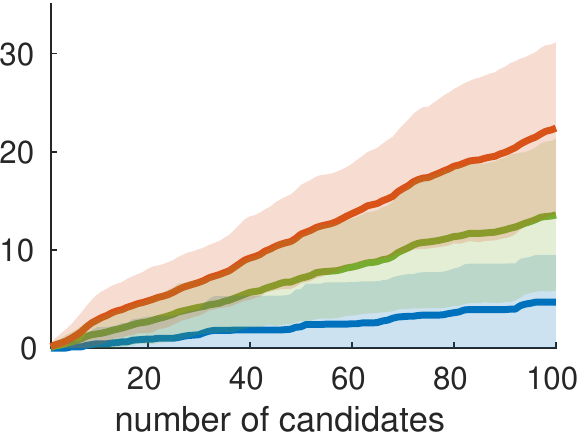}
    \vspace{-14pt}
    \caption{\footnotesize{Number of possibly safe (green) and undecided (orange) determined by the method. The remainder is determined certainly unsafe. In blue is the number of safe candidates with respect to the true environment. Shaded area is one standard deviation.}}
    \vspace{0pt}
    \label{fig:suite_planning}
\end{wrapfigure}

For a given demonstration (see Figure \ref{fig:cloud_planning}), we sample 100 candidates using hit-and-run \cite{hit_and_run} and evaluate each for safety by solving Prob. \ref{prob:cand_traj}. We run this experiment 50 times.
The classification of 
safe paths results in an average of $13.56 \pm 7.31$ possibly safe, $22.40 \pm 8.84$ undecided, and $64.04 \pm 13.02$ certainly unsafe (see Fig \ref{fig:suite_planning}). 
Of the possibly safe paths, an average of $4.68 \pm 4.82$ are safe with respect to the true environment. With regards to running time, we limit execution  of the CSP to 1 second. Generating trajectories took an average of $0.1405$ seconds per trajectory and finding consistent environment took $1.9692$ seconds per candidate, totaling $2.1097$ seconds.

Since we assume the robot acts with kinematic constraints, we can execute a potentially safe path and backtrack if we discover it is not safe. If there exists at least one possibly safe path which is truly safe, then we will safely reach the goal with this strategy. In Fig. \ref{fig:cloud_planning}, note the portion of the dashed black and orange paths near the demonstration vertex renders the black candidate certainly unsafe which highlights that our algorithm is capable of correctly classifying two similar paths.

%\begin{wrapfigure}{r}{0.35\textwidth}
%    \vspace{-35pt}
%    \includegraphics[width=0.35\textwidth]{figures/scout.pdf}
%    \vspace{-20pt}
%    \caption{\footnotesize{The true environment (purple) for a scouting robot with intermediate goals $x_{G_1}$ and $x_{G_2}$. Note the true environment violates the assumptions as the obstacles are enclosed. Each sample of the reconstructed environments is plotted transparently in red. Here the CSP found 5 assignments, and we sampled 100 obstacles from each.}}
%    \vspace{0pt}
%    \label{fig:scout_results}
%\end{wrapfigure}

\vspace{-8pt}
\subsection{Sequential Demonstrations and Prob- \\ abilistic Environment Representation}

% \begin{figure}[t]
%     \centering
%     \vspace{-10pt}
%     \begin{minipage}{0.35\textwidth}
%         \centering
%         \includegraphics[width=\textwidth]{figures/scout.pdf}\\
%         (a)
%     \end{minipage}%
%     \hspace{15pt}
%     \begin{minipage}{0.44\textwidth}
%         \centering
%         \includegraphics[width=\textwidth]{figures/sampling_close.pdf}\\
%         (b)
%     \end{minipage}
%     \vspace{-8pt}
%     \caption{\footnotesize{(a) The true environment (purple) for a scouting robot with intermediate goals $x_{G_1}$ and $x_{G_2}$. Note the true environment violates the assumptions as the obstacles are enclosed. 300 sample obstacles from 3 assignments are transparently plotted in red. (b) Distance of true obstacle to the cloud as samples are taken. The magnitude of the distance depends on the scale of the obstacle, however, a downward trend in distance is apparent as more samples are drawn. The resulting samples, true obstacles, and demonstration are shown above the corresponding range for each assignment. Each dotted line signifies a change from sampling from one assignment to another.}}
%     \vspace{20pt}
%     \label{fig:sampling}
% \end{figure}

\begin{figure}[t]
    \centering
    \vspace{-12pt}
    \begin{minipage}{0.3\textwidth}
        \centering
        (a)\includegraphics[width=\textwidth]{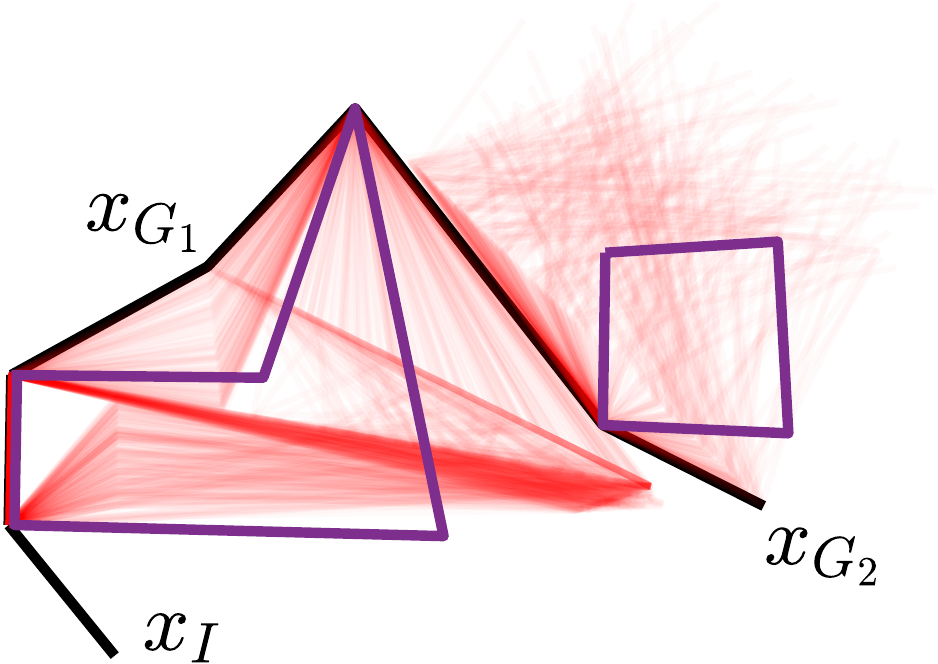}
    \end{minipage}%
    \hspace{40pt}
    \begin{minipage}{0.4\textwidth}
        \centering
        \includegraphics[width=\textwidth]{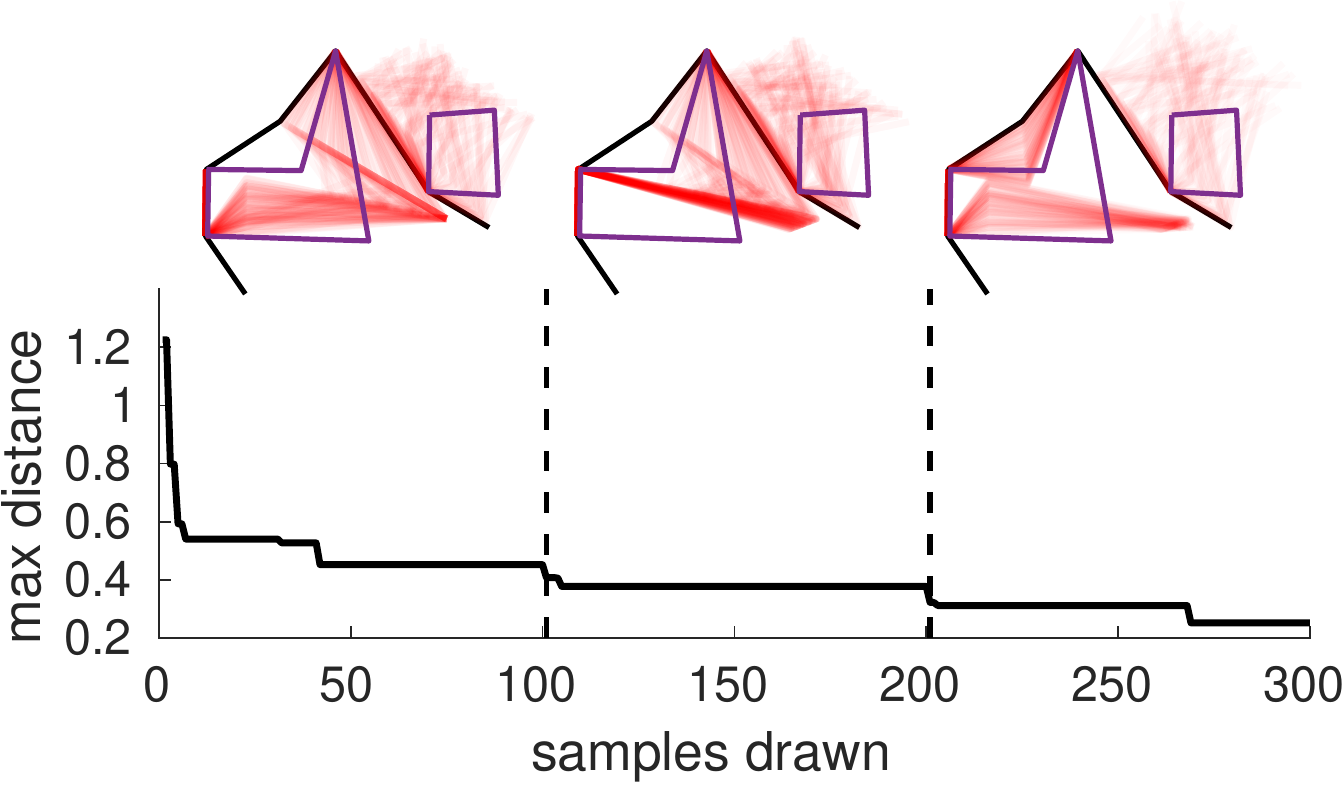}(b)
    \end{minipage}\\
    \vspace{-8pt}
    \begin{minipage}{0.28\textwidth}
        \centering
        (c)\includegraphics[width=\textwidth]{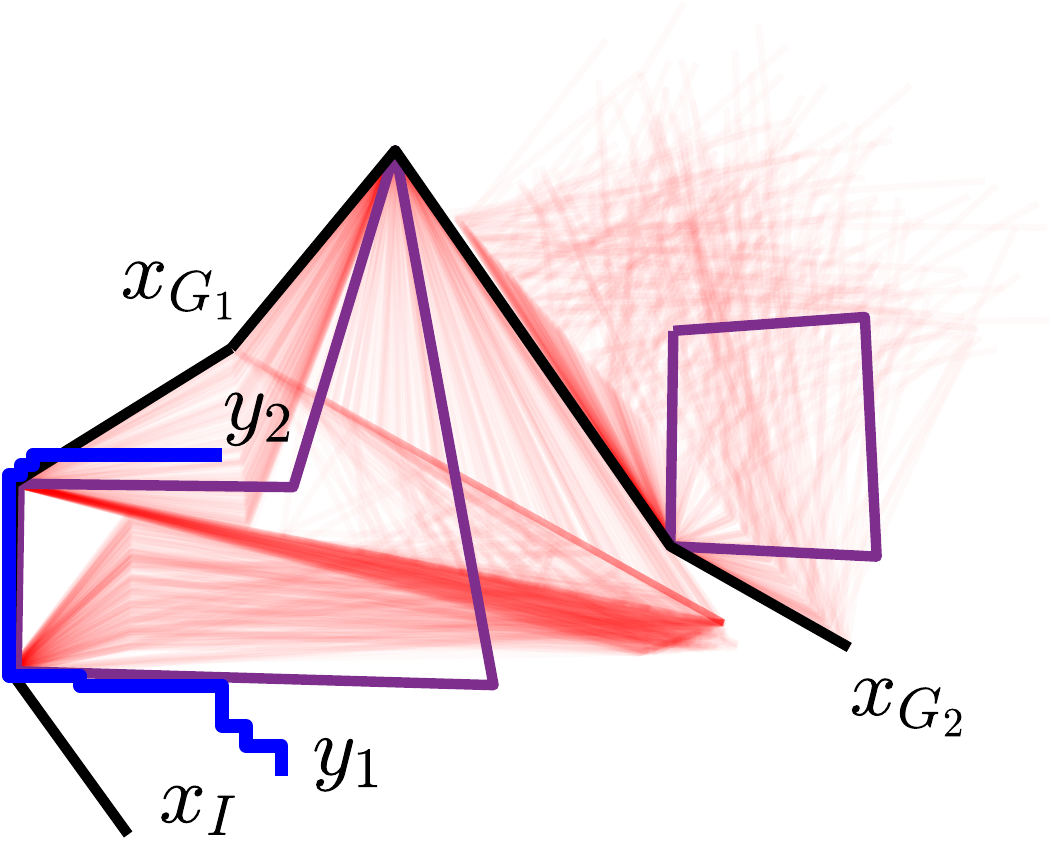}
    \end{minipage}%
    \hspace{40pt}
    \begin{minipage}{0.45\textwidth}
        \centering
        \vspace{23pt}
        \includegraphics[width=\textwidth]{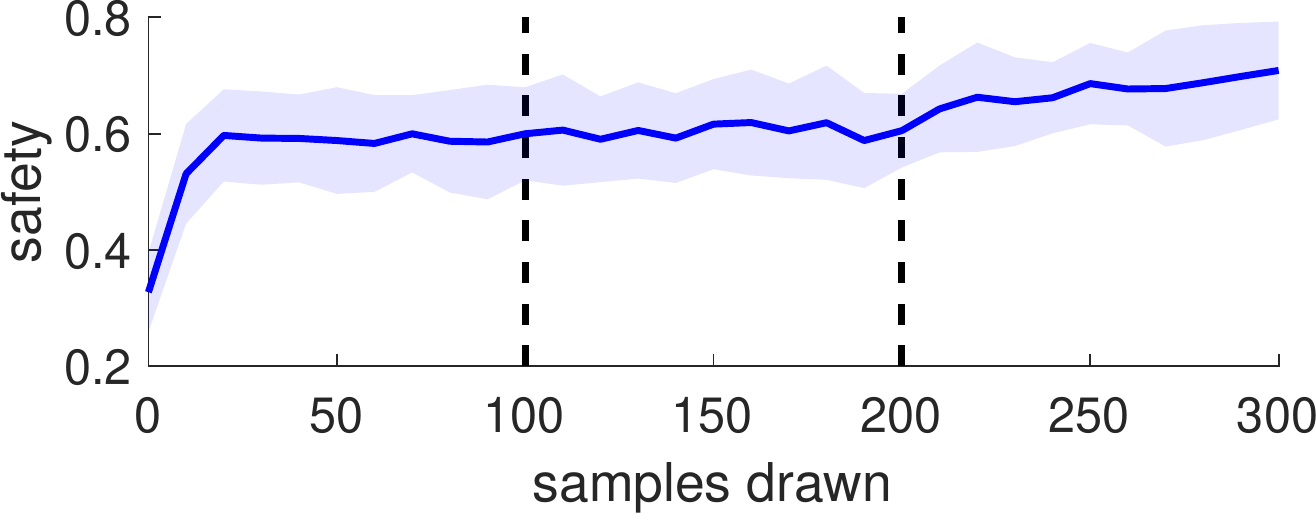}(d)
    \end{minipage}
    \vspace{-8pt}
    \caption{\footnotesize{(a) The true environment (purple) for a scouting robot with intermediate goals $x_{G_1}$ and $x_{G_2}$. Note the true environment violates the assumptions as the obstacles are enclosed. 300 sample obstacles from 3 assignments are transparently plotted in red. (b) Distance of true obstacle to the cloud. The magnitude of the distance depends on the scale of the obstacle, however, a downward trend in distance is apparent as more samples are drawn. The resulting samples, true obstacles, and demonstration are shown above the corresponding range for each assignment. A dotted line indicates when samples are taken from a new assignment. (c) A safe example plan (blue) from $y_1$ to $y_2$ planned with MCR (d) Estimated probability of safety of planning with MCR for random start and goal with 3-$\sigma$ bound.}}
    \vspace{18pt}
    \label{fig:plan}
\end{figure}

In Fig. \ref{fig:plan}a, we show the results of reconstructing 
the environment from the sequential demonstrations of a scouting robot. We ran the CSP for 10 hours and then used the resulting assignments for recovering obstacles if the assignments produced a configuration $O$ that solved Prob. \ref{prob:env_prob}. In this case, the first assignment was found in 20 minutes and the next 3 in 2 hours. We sampled 100 obstacles from the 5 assignments found, taking 2 additional minutes. We only sampled consistent obstacles from 3 of the 5 assignments.

%\begin{wrapfigure}{r}{0.5\textwidth}
%    \centering
%    \vspace{-25pt}
%    \includegraphics[width=0.45\textwidth]{figures/sampling_close.pdf}
%    \vspace{-10pt}
%    \caption{\footnotesize{Distance of true obstacle to the cloud as samples are taken. The magnitude of the distance depends on the scale of the obstacle, however, a downward trend in distance is apparent as more samples are drawn. A hundred samples are drawn from each assignment, and the resulting samples, true obstacles, and demonstration are shown above the corresponding range. Each dotted line signifies a change from sampling from one assignment to another.}}
%    \label{fig:distance}
%\end{wrapfigure}

To evaluate success, we measure the distance of the true obstacles $\mathbb{O}_\textrm{true}$ to the sample cloud $\mathbb{O}$ as samples are taken. Let the distance be $d(\mathbb{O}_\textrm{true}, \mathbb{O}) \doteq \textrm{max}_{o \in \mathbb{O}_\textrm{true}} \, \textrm{min}_{o' \in \mathbb{O}} \, || o - o' ||_2$, approximated by finely subsampling the cloud and true obstacle (see Fig. \ref{fig:plan}b). This metric is specific to the given environment, but the maximum distance asymptotically trends towards a lower bound for each assignment and sampling from a new assignment further decreases the distance.

We can use this obstacle cloud to also plan new paths that maximize the likelihood of safety. If we interpret each sample as equally likely, this path planning problem reduces to Minimum Constraint Removal (MCR) \cite{hauser2014minimum}. After a batch of 10 samples are taken, we discretize the space and plan a path using (MCR) for 5000 randomly selected start and goal configurations within the bounding box of the environment (see Fig. \ref{fig:plan}cd). We interpret the fraction of planned paths that do not intersect with the true obstacle as the estimated probability of safety. We see that safety improves with only a few samples and continues to improve with more samples and new assignments. Drops in safety are due to random sampling of the start and goal as well as obstacle samples.
%either a result of unlucky start and goal sampling or drawing obstacle samples that do not align with the true obstacle.

%\begin{wrapfigure}{r}{0.4\textwidth}
%    \centering
%    \includegraphics[width=0.4\textwidth]{figures/cloud_plan.pdf}
%    \caption{A safe example plan (blue) from $y_1$ to $y_2$ in the presence of an obstacle cloud (red). The true obstacle is shown in purple.}
%    \label{fig:cloud_plan}
%\end{wrapfigure}

%\begin{wrapfigure}{r}{0.4\textwidth}
%    \centering
%    \vspace{-30pt}
%    \includegraphics[width=0.4\textwidth]{figures/safe_planning.pdf}
%    \vspace{-20pt}
%    \caption{\footnotesize{Estimated probability of safety as samples are drawn. The dotted lines indicate when samples are taken from a new assignment.}}
%    \label{fig:safe_planning}
%\end{wrapfigure}

This environment is challenging since the demonstrator navigates around one obstacle in both demonstrations.
%Determining the environment from each demonstration individually could produce an overall inconsistent environment, as obstacles found in the first demonstration may occlude obstacles in the second. 
Our method infers the knowledge gained from previous demonstrations in reconstructing obstacles in future demonstrations.

\vspace{-12pt}
\section{Conclusion}
\vspace{-5pt}

In this paper we present two methods for the novel problem of inferring obstacles and path validity from visibility-constrained demonstrations. The first is probabilistically complete in finding obstacle configurations that solve Prob. \ref{prob:cand_traj}, but is intractable to execute. The second is a tractable method that is capable of finding a subset of the possible true environment configurations and in some cases can assert the candidate path is unsafe. This method can be applied to planning problems or environment reconstruction either online or offline.

% future work stuff can be shortened/cut to save space
%Future lines of work for this research include scoring each obstacle according to some ``energy" function \cite{Chou2018LearningConstraints} so that we can weigh obstacles according to their energy. 
%Furthermore, it may be possible to accommodate other vision models (for example limited range) or demonstrator strategies. 
%Perhaps more importantly, most demonstrations are suboptimal whereas we assume the demonstrator is optimal with respect to the visibility constraints. Handling $\delta$-suboptimal \cite{Chou2018LearningConstraints} visibility-constrained solutions to Prob. \ref{prob:partially_optimal} would ease application to demonstrations given by human beings. Additionally, handling demonstrations from different demonstrators could improve the learnability of the environment.

This work could be improved upon by extending to obstacles that are enclosed, can intersect or are free-standing (potentially by searching for subgraphs in the CSP) will allow generalization to many environments. Furthermore, extension to multiple demonstrations is possible as long as obstacles can be sampled in a way that satisfies visibility constraints for both demonstrators. Additionally, it may be possible to use the structure of the demonstration (connected line segments) to extend to other vision models such as a forward facing cone. Moreover, assuming the demonstrator optimally plans paths can be restrictive when using human demonstrations; we can address this by modifying verification to partially align inferred plans with the demonstration.
%Regardless, improving running time, perhaps through a more informed search in the CSP and especially with regards to an increased number of survey points, is critical to addressing the scalability of this method.

\vspace{-10pt}

\bibliographystyle{IEEEtran}
\bibliography{references.bib}

\newpage

\appendix
\section{Proof of Theorem \ref{th:exit_point}}
\label{app:th1_proof}

Before restating and proving Thm. \ref{th:exit_point}, we clarify the mathematical subtlety present in Prob. \ref{prob:partially_optimal}. Prob \ref{prob:partially_optimal} assumes the existence of a path with minimum cost. Recall obstacles are line segments and consider a path that goes around an obstacle to get to the goal. The point nearest to the obstacle vertex lies some nonzero distance away from the obstacle. For such a path, we can produce a lower cost path that lies even closer to the obstacle. Therefore a minimum is not guaranteed to exist.

To mitigate this problem, let $\Xi_{t,\textrm{safe}} = \{\xi \in \Xi \,|\, \xi(\tau) \cap O(t) = \emptyset \enspace \forall \tau \in (0,1]\}$. We allow the start of the path to intersect an obstacle for a reason that will be clear momentarily. Then, consider the closure of $\Xi_{t,\textrm{safe}}$ denoted $\textrm{cl}(\Xi_{t,\textrm{safe}})$. To be precise, a path $\xi$ lies in $\textrm{cl}(\Xi_{t,\textrm{safe}})$ if for all $\epsilon > 0$ there exists $\xi'$ such that $\xi' \,\cap\, O(t) = \emptyset$ and $F(\xi,\xi') < \epsilon$ where $F(\cdot,\cdot)$ denotes Fr\'{e}chet distance. This closure contains paths that intersect with the known obstacle set, but do not cross over the obstacles. We can redefine Problem \ref{prob:partially_optimal} as

\begin{problem}[Demonstrator's Planning Problem]\label{prob:partially_optimal_alt}
\begin{equation}\label{eq:partially_optimal_alt}
    \begin{split}
        \begin{aligned}
            \plan\xi \enspace = \quad& \text{arg min}_{\xi}
            & & c(\xi) \\
            & \text{subject to}
            & & \xi(0) = a, \quad \xi(1) = b \\
            & & & \xi \in \textrm{cl}(\Xi_{t,\textrm{safe}})
        \end{aligned}
    \end{split}
\end{equation}
\end{problem}

Now the plan always exists as the minimum exists in the closure of this set. However, when replanning the demonstrator may lie upon the vertex of an obstacle. We allow plans to start in the obstacle set, and furthermore restrict the entire demonstration to lie in $\textrm{cl}(\Xi_{\textrm{safe}}) = \textrm{cl}(\{\xi \in \Xi \,|\, \xi(\tau) \cap O = \emptyset \forall \tau \in [0,1]\})$. We can then rewrite Prob. \ref{prob:arrive_at_goal} as

\begin{problem}[Demonstrator's Strategy]\label{prob:arrive_at_goal_alt}
Find $\demo:[0,1] \rightarrow \X$ such that $\demo(0) = x_I$, $\demo(1) = x_G$, $\demo \in \textrm{cl}(\Xi_{\textrm{safe}})$ and is generated with the following strategy. At any time $t$ with the demonstrator at position $\demo(t)$, the demonstrator follows $\plan\xi$, the solution to Prob. \ref{prob:partially_optimal} for start $\demo(t)$, goal $x_G$, and known obstacles $O(t)$, i.e. $\exists \delta > 0, u > 0$ such that $\xi^*(t + \frac{\tau}{u}) = \plan\xi(\tau)$ $\forall \tau \in [0, \delta)$.
\end{problem}

As a result, each plan $\plan\xi$ carries an additional constraint that the resulting demonstration lies in the closure of the safe set of paths (i.e. does not cross over obstacle line segments). Now that intermediate plans and demonstrations are well-defined, we move on to proving Thm. 
\ref{th:exit_point}.

\textit{Theorem \ref{th:exit_point}:}
At time $t$, suppose $\plan\xi$ solves Prob. \ref{prob:partially_optimal}, $\plan\xi$ is not a straight line to goal, and $x^* = \plan\xi(s)$ is the first point on $\plan\xi$ coincident with an obstacle vertex. Then, $\demo(t+\tau) = \plan\xi(\tau)$ for all $\tau \in [0, s]$. Therefore, the demonstrator will never deviate from a plan unless at an obstacle vertex.

\begin{proof} For the following argument, we refer to $\plan\xi$ as $\plan\xi_1$. Suppose at time $t$, we are at position $\demo(t)$ and $\plan\xi_1$ solves Problem \ref{prob:partially_optimal}, for start $\demo(t)$, goal $x_G$, and known obstacles $O(t)$.

Now suppose at time $s_2$, $t < s_2 < s$, $\plan\xi_2$ solves Prob. \ref{prob:partially_optimal} with start $\demo(s_2)$ and let $y^* = \plan\xi_2(s_3) \neq x^*$ be the first point on $\plan\xi_2$ coincident with an obstacle vertex. If the plan does not have a point coincident with an obstacle vertex then it is either not optimal or it is a line straight to goal. We assume the demonstrator always plans optimal paths with respect to known knowledge, and additionally if the plan goes directly to goal the demonstrator will never deviate. Recall the plan will also be constructed of consecutive line segments (see Visibility Graphs in \cite{Lav06}). Let time $s_2$ be the first time such a plan exists that deviates from the plan $\plan\xi_1$. See Fig. \ref{fig:theorem1} for a visualization of the situation described.

\begin{figure}
    \centering
    \vspace{7pt}
    \includegraphics[width=0.6\textwidth]{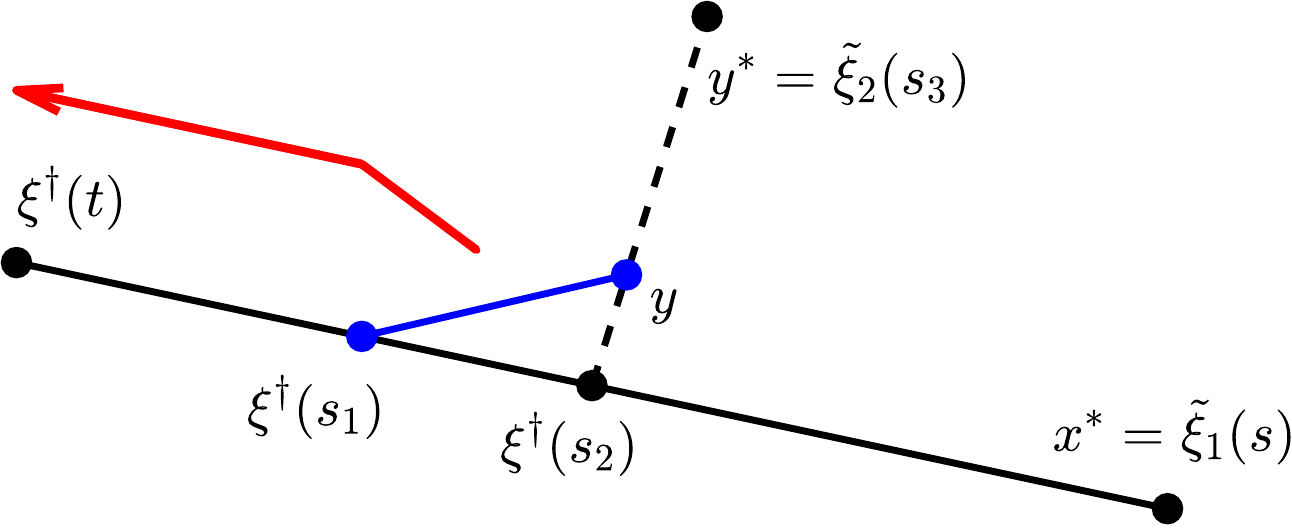}
    \caption{\small{Visualization of the proof of Theorem \ref{th:exit_point}. Here the red line is an obstacle that blocks the point $y^*$ from $\demo(t)$. The blue line indicates the line of sight between $\demo(s_1)$ and $y$. The black line corresponds to the initial plan $\plan\xi_1$ and the dotted line is the alternative plan $\plan\xi_2$ that is planned before reaching $x^*$. There must exist some point $y$ that is visible before reaching the point $\demo(s_2)$. If $y$ is visible at $\demo(s_1)$, then a shorter path can be constructed that goes from $\demo(s_1)$ to $y$ and then to $y^*$. This is a contradiction. Therefore, when the demonstrator is at point $\demo(t)$, they will not deviate from the plan until they reach the point $x^* = \plan\xi_1(s)$.}}
    \label{fig:theorem1}
\end{figure}

Suppose there does not exist some point $y$, on the line segment $(\demo(s_2), \plan\xi_2(s_3))$, that is visible from some point on the demonstration $\demo(s_1)$. For such a point to not exist, there must exist some obstacle $o$ (or multiple) such that $o \cap (\demo(\tau_1), \plan\xi_2(\tau_2)) \neq \emptyset \enspace \forall \tau_1 \in [t, s_2), \tau_2 \in (0, s_3]$. Clearly, such an obstacle only exists if the obstacle has a vertex coincident with the point $\demo(s_2)$. Therefore, there always exists a point $y$ that is visible on the line segment $(\demo(s_2), \plan\xi_2(s_3))$ from some point $\demo(s_1)$.

If such a point is visible, we can construct a shorter path from $\demo(s_1)$ to $y$ and then to $y^*$. Such a path is clearly shorter by virtue of the triangle inequality. However, we assumed $s_2$ was the first time that the demonstrator would plan a lower cost path which is a contradiction, so no better path $\plan\xi_2$ can be discovered until the demonstrator reaches the point $x^*$.

\end{proof}

\section{Cell Sequence Identification Example}\label{app:csqid}

\begin{figure}
    \centering
    \includegraphics[width=0.6\textwidth]{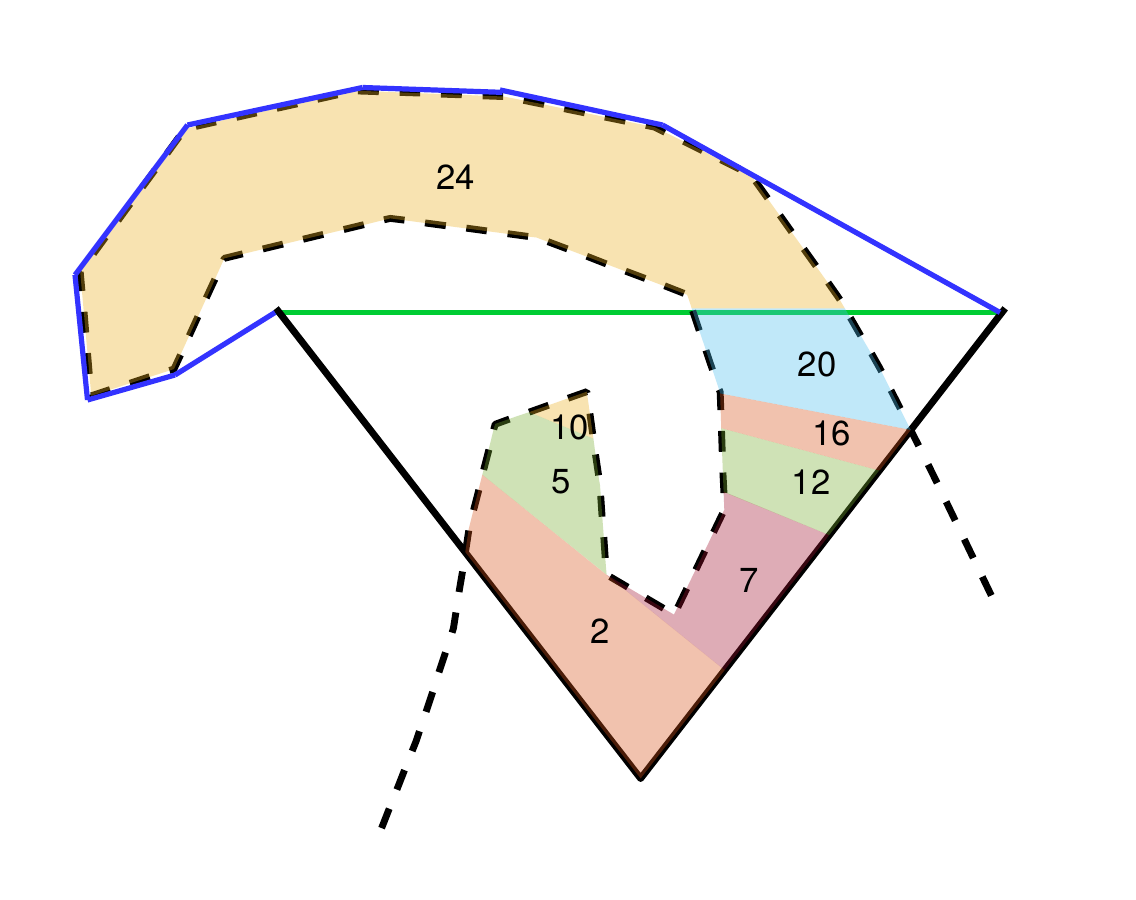}
    \caption{Example of cell sequence identification. The demonstration is solid black, the candidate path is in dashed black, and select cells are shown in varying colors. The green path is the shortest alternative path for the obstacle $o$. The blue path is the shortest alternative for the obstacle $o'$. For both configurations of environments, the demonstration is the same as the shortest inferred path.}
    \label{fig:csqid}
\end{figure}

Here we walk through an example of cell sequence identification. Consider the demonstration and candidate in Figure \ref{fig:csqid}. First we note that an obstacle must have a vertex coincident with the intermediate vertex of the demonstration by Theorem 1. Furthermore, the obstacle must lie in cell 2 (it does not start below the demonstration) otherwise the demonstration would not be optimal. The only cells of interest are those that are enclosed by the demonstration and the candidate ($\{\cell_2, \cell_5, \cell_7 ,\cell_{10} , \cell_{12}, \cell_{16}, \cell_{20}, \cell_{24}\}$). Let's consider two cell sequences.

\begin{equation}
    \class = \{\cell_2, \cell_5, \cell_{10}\}, \quad \class' = \{\cell_2, \cell_7, \cell_{12}, \cell_{16}, \cell_{20}, \cell_{24}\}
\end{equation}

Consider the obstacles that fills the entirety of the cells in each cell sequence. For $o = (\cell_2 \cup \cell_5 \cup \cell_{10}) \setminus \safeset$ and $o' = (\cell_2 \cup \cell_7 \cup \cell_{12} \cup \cell_{16} \cup \cell_{20} \cup \cell_{24}) \setminus \safeset$, there are two homology classes of interest, one where paths go above the obstacle and one where paths go below the obstacle. In both cases, the shortest path that goes below the obstacle is exactly the demonstration and so this is the inferred robot path. The shortest paths (shown in green and blue for $o$ and $o'$) that go above the obstacle are the alternative robot paths. We emphasize that both of these paths are found without considering visibility constraints i.e. the known obstacle set is the entire obstacle.

For the obstacle $o$, the green path is shorter than the demonstration so $\bar{c}^1_a < \underline{c}^1_\textrm{inf}$. This implies that there does not exist any obstacle that lies inside the cell sequence $\class$ that will produce a consistent environment. However, for $o'$ the blue path is longer than the demonstration and $\bar{c}^1_a \geq \underline{c}^1_\textrm{inf}$. Since the alternative robot path is longer than the inferred robot path, we know that there may exist an obstacle that also satisfies visibility constraints. In this class, the cell sequence identification procedure would produce a singleton set $\classid = \{\class'\}$ which vastly simplifies the workload for the CSP, indeed, it will simply return $\class'$ as the only viable assignment.

\section{Obstacle Recovery}\label{app:obs_rec}

We seek to find $m \leq n$ obstacles in $\X$ (curves parameterized by line segments) from the assignment $\assign$. We will first chain survey-cell paths if necessary, producing sets of survey-cell paths. Consider two survey-cell paths $\graphpath_i$ and $\graphpath_j$ with a nonempty overlap. If $\chainable{\graphpath_i}{\graphpath_j}$ is true and there doesn't exist two obstacle curves that don't occlude one another, then we replace $\graphpath_i$ and $\graphpath_j$ with a new path $\graphpathvar$ that is formed by combining $\graphpath_i$ and $\graphpath_j$. Otherwise, we can choose whether or not to combine. For instance, recall the example in the previous section of two survey-cell paths overlapping only at the terminal node. There could be a gap between the two, or they could be combined into one obstacle curve. In this case, we create two resulting assignments, one where the survey-cell paths are chained and one where they are not. We repeat this process for each pair that may or may not be chained.

There is a special case here that must be considered before reconstructing obstacle curves. Though we enforce that each pair of survey-cell paths are chainable, this does not enforce that the entire assignment will reduce to a set of valid survey-cell paths. For instance, consider the following 3 paths (with numbers representing unique nodes): $1 \rightarrow 2$, $3 \rightarrow 2$, and $4 \rightarrow 2$. All three are pairwise chainable, but they cannot all be reduced to a single path. If we cannot reduce the paths, then we return to the CSP to find a new assignment. 

%Suppose we require some subset of survey-cell paths to be chained. We also enforce that the resulting $n$ survey-cell paths $(\graphpath_1, ..., \graphpath_n)$ reduce to a set of \textit{compatible} $m$ survey-cell paths $(\rho_1, ..., \rho_m)$ if each survey-cell path $\graphpath_i$ (or its reverse) is a subpath of some $\rho_j$, each $\rho_j$ is formed by combining $\graphpath_i$'s together by overlapping either their prefix or suffix, and no $\rho_j$ shares a node with any other $\rho_k$. In order to satisfy Assumptions \ref{assume:line_segments} - \ref{assume:visible}, we require that each assignment can be reduced to a compatible set of survey-cell paths. For shorthand, we say $\mathcal{C}(\assign)$ is true if the assignment $\assign$ is reducible.

For an assignment $(\graphpathvar_1, \ldots, \graphpathvar_m)$ we seek to find corresponding obstacle curves that are consistent with the demonstration. We partition each survey-cell path $\graphpathvar_i$ into a minimum number of nonoverlapping survey-cell paths $\seg_{i,1}, \ldots, \seg_{i,\numseg}$ such that each path is only seen from one survey point.
%where $\numseg$ is minimized, satisfying the following two conditions for $k \in \{1, \ldots, \numseg\}$:
%\vspace{-13pt}
%\begin{equation}
%            \hspace{-2pt}\left(\graphpath_i = \seg^i_1 \concat \ldots \concat \seg^i_{\numseg}\right) \bigwedge
%            \left(\exists \alpha_j \enspace \text{s.t.} \enspace \{\alpha_j\} = \image{\nodesp{(\seg^i_{k})}}\right)\hspace{-10pt}
%\end{equation}
%\vspace{-13pt}
Intuitively, we are separating an obstacle into the different parts seen by different survey points (see Figure \ref{fig:representation}). In this section, we will refer to an obstacle segment $\obsseg_{i,k} \subset \X$ as the realization of a survey-cell segment $\seg_{i,k}$ of survey-cell path $\graphpathvar_i$. Then, the obstacle curve is precisely $o_i = \bigcup_{k=1}^{\numseg} \obsseg_{i,k}$.

Neighboring obstacle segments $\obsseg_{i,k}$ and $\obsseg_{i,k+1}$ (terminating at cell $\cell_1$ and starting at cell $\cell_2$, respectively) are constrained to intersect at a \textit{join point} denoted $\joinpt_{k,k+1}$ (yellow line in Figure \ref{fig:representation}). We choose $\joinpt_{k,k+1} \in \left(\cell_1 \cap \cell_2\right) \setminus \safeset$, i.e. the join point lies in the shared region of the two survey-cell segments%
%or in the overlapping cell if $\nodecell{(\seg^j_l(\length{v^j_l})} = \nodecell{(\seg^j_{l+1}(1))}$ %space saver
. This ensures the obstacle can be constructed with a continuous curve in $\X$. Suppose $\graphpathvar_i$ is constructed from multiple overlapping survey-cell paths. Each of these underlying survey-cell paths originate from an intermediate vertex of the demonstration, all together denoted $x_{\mathcal{J}}$, $\mathcal{J} \subset \{2, \ldots, n+1\}$. We require $o_i$ to intersect each start point $x_{j}$, $j \in \mathcal{J}$, defined as a \textit{point constraint}. 

With the given join points and point constraints, we aim to construct an obstacle that does not self occlude to ensure each obstacle segment is visible to the corresponding survey point. We can construct segments $\obsseg_{i,k}$ beginning and ending at points $a$ and $b$ by sampling continuous radial functions with origin at $\nodesp{\seg_{i,k}}$, defined as follows. Suppose $a$ and $b$ lie at angles $\phi_1$ and $\phi_2$ with respect to $\nodesp{\seg_{i,k}}$. Then a radial function $f_{i,k} : [\phi_1, \phi_2] \to [0, \infty)$ maps angle to radius, producing a curve $(f_{i,k}(\phi), \phi)$ in polar coordinates. We restrict this obstacle segment $\obsseg_{i,k}$ to lie in $\classobs{\nodecell{\seg_{i,k}}}$, visit any point constraint necessary, and be constructed with line segments. This obstacle curve is clearly not self-occluding from the survey point, otherwise it would not be a radial function. We do not discuss a strategy to sample general radial functions but show an exact method in Sec. \ref{sec:convex_grad} (also see Fig. \ref{fig:segment_combo}).

\section{Proof of Theorem \ref{th:prob_comp}}\label{app:prob_comp_proof}

\textit{Theorem \ref{th:prob_comp}:} 
Let $\mathbb{O}_{n_1}$ be the set of samples drawn by the $n_1^{th}$ iteration of the outer loop of Alg. \ref{alg:m1_find_environment_conf}. Then, under Asm. \ref{assume:line_segments} - \ref{assume:n_curves} and for some $0 < K < \infty$ and $0 < \epsilon_2 < 1$,

\vspace{-8pt}
\begin{equation}
\mathbb{P}(\mathbb{O}_{n_1} \cap \solnenv \neq \emptyset) \geq 1 - (1-\epsilon_2)^{n_1 - K} \qquad \forall n_1 \geq K
\end{equation}

Before stating the proof, we add a small technical assumption to the sampling method. We place a restriction on sampling obstacle segments such that sampling vertices in a set with nonzero area has nonzero probability. 

\begin{proof}
Each $\env' \in \solnenv$ consists of $m \leq n$ obstacle curves each starting at some $x_i$ by Assumption \ref{assume:n_curves}. Since only the portion of the obstacle visible to the demonstrator by the third to last vertex is learnable, we show completeness in reconstructing the obstacle $O'' = O' \cap R(t_n)$.

Since $\env'' \subset R(t_n)$, we can partition each curve into obstacle segments where each obstacle segment is seen from a survey point $\spt$ and the segment is a radial function with origin at $\spt$. Each of these segments lie entirely in a cell sequence and therefore can be represented with a survey-cell path in $G$, resulting in $m$ survey-cell paths $\graphpathvar_1, \ldots, \graphpathvar_m$. 

The CSP finds $n \geq m$ paths, so we potentially need to augment the set of paths to find a corresponding assignment. For each $x_i$, $i \in \{2, \ldots, n+1\}$ that does not coincide with the start of an obstacle curve, $x_i$ must still be coincident with an obstacle vertex on the curve $o_\kappa$ for some $\kappa \in \{1, \ldots, m\}$, say corresponding to survey-cell path $\graphpathvar_\kappa$ and node $\node = \graphpathvar_\kappa(k)$ for some $k$. We then augment the set of survey-cell paths with a survey-cell subpath $\graphpathvar$ which starts at a node with corresponding cell bordering $x_i$ and terminates at the end of the path. Intuitively, we are constructing a survey-cell path that overlaps an existing survey-cell path with a corresponding obstacle curve that starts from an intermediate vertex. This produces $n$ total survey-cell paths resulting in $\Graphpath = \{\graphpath_2, \ldots, \graphpath_{n+1}\}$ which we order by corresponding intermediate vertex, i.e. $\graphpath_i$ starts at vertex $x_{i}$.

Now we seek to show there exists an assignment of the CSP corresponding to the survey-cell paths $\{\graphpath_1, \ldots, \graphpath_n\}$ satisfying the constraints. Let's consider each constraint individually.

\eqref{eq:constraint_start} enforces that each survey-cell path $\graphpath_i$ starts at a cell containing $x_{i}$. By Assumption \ref{assume:n_curves} and the construction above, this is trivially satisfied.

By Section \ref{sec:verifying_conf}, it is clear that in order for $\env$ or $\env' \in \solnenv$ to be solutions, the cost of alternative paths must be greater than the respective inferred paths. Therefore, $[\nodecell\graphpath_1, \ldots, \nodecell\graphpath_{n+1}] \in \classid$ and constraint \eqref{eq:constraint_class_id} is satisfied.

For \eqref{eq:constraint_vision_complete}, let $\mathcal{N}_{\textrm{occ}}$ be the set of all nodes fully occluded by the assignment $\Graphpath$. If any $\graphpathvar \in \Graphpath$ contains any $\node \in \mathcal{N}_{\textrm{occ}}$ then the corresponding obstacle segment is not visible to $\nodesp\node$. This contradicts the fact that $\env'' \subset R(t_n)$. Therefore, we do not eliminate $\Graphpath$ with constraint \eqref{eq:constraint_vision_complete}.

For constraint \eqref{eq:constraint_overlapping_complete}, consider two paths $\graphpath$ and $\graphpath'$ that do not chain (otherwise the constraint is trivially satisfied). These two paths may share a node $\node$. However, if they do then the corresponding obstacle $o$ and $o'$ certainly do not occlude one another by virtue of the construction of the paths. All that remains to be shown is that the obstacles do not intersect one another. If they intersect, they must be connected to form a single curve by the assumption that the environment contains no intersecting obstacles. If they are connected in a way that forms a single curve, then our construction of the paths would have formed a single path for the entirety of the obstacle. Then, either $\graphpath$ or $\graphpath'$ must have been constructed using a subsequence of the nodes of the other and the two paths chain. This is a contradiction, so we conclude that constraint \eqref{eq:constraint_overlapping_complete} does not eliminate the assignment $\Graphpath$.Therefore none of the constraints would prevent the assignment of $\Graphpath$ by the CSP. 

Now we show that the full method has nonzero probability of sampling $\env''$. First we note each corresponding obstacle segment $\obsseg_{\kappa,l}$ consists of $k_{\kappa,l}$ line segments by Assumption \ref{assume:line_segments}. Define $K^{\textrm{soln}}_2 \doteq \max_{\kappa,l} k_{\kappa,l}$ and $K^{\textrm{soln}}_1 \doteq \max_{\graphpathvar_\kappa} \length{\graphpathvar_\kappa}$ where $\length{\cdot}$ is the length of the survey-cell path. There is a segment discretization $\segdisc$ corresponding to all $\{\env' \cap R(t_n) \mid \env' \in \envset\}$. Once our algorithm loops to the point $K_1 \geq K^{\textrm{soln}}_1$, $K_2 \geq K^{\textrm{soln}}_2$, at each step we have $\epsilon_2 > 0$ chance of sampling from the set $\{\env' \cap R(t_n) \mid \env' \in \envset\}$ (provided our sampler has a nonzero likelihood of sampling vertices from a non-zero measure set). Let $K = \max(K^{\textrm{soln}}_1, K^{\textrm{soln}}_2)$. Then for all $n_1 \geq K$,

\begin{equation}
\mathbb{P}(\mathbb{O}_{n+1} \cap \solnenv \neq \emptyset) \geq 1 - (1-\epsilon_2)^{n_1 - K}
\end{equation}

\end{proof}

\section{Proof of Theorem \ref{th:cand_unsafe}}\label{app:cand_unsafe}

\textit{Theorem \ref{th:cand_unsafe}:}
Suppose Assumption \ref{assume:line_segments} and \ref{assume:n_curves} is satisfied and $\classid = \emptyset$. Then the candidate robot path $\candtraj$ is unsafe.

\begin{proof}
Suppose there exists an environment configuration that is consistent with the demonstration and renders the candidate trajectory $\candtraj$ safe. Let $\mathbf{O}$ be a union of the cells such that the environment configuration $O \subseteq \mathbf{O}$. In order for $O$ to be consistent, $c(\plan\xi^i_{\textrm{inf}}) \leq c(\plan\xi^i_a)$ for all $i \in \{1, \ldots, n\}$. Since $\underline{c}^i_{\textrm{inf}}$ is a lower bound on the cost of the inferred path and $\bar{c}^i_a$ is an upper bound on the cost of the alternative for any $O \in \mathbf{O}$, we have $\bar{c}^i_a \geq \underline{c}^i_{\textrm{inf}}$ for some set of cells. This is a contradiction, so there must not exist a consistent environment configuration.
\end{proof}

\section{Specialized Obstacle Recovery} \label{app:special_obs_rec}

\begin{figure}[ht]
    \centering
    \includegraphics[width=0.4\textwidth]{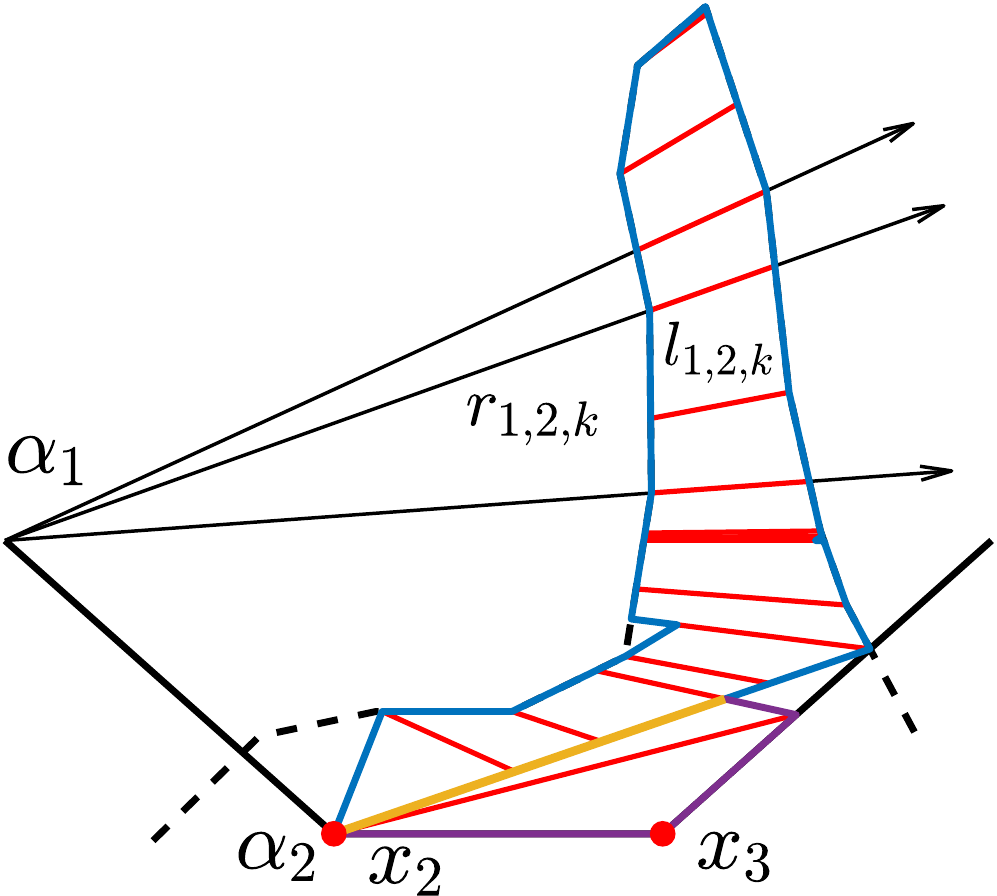}
    \caption{\footnotesize{Ray based convex decomposition of two cell sequences (purple and blue). See Figure \ref{fig:representation}.}}
    \vspace{20pt}
    \label{fig:segment_combo}
\end{figure}

For each survey-cell segment $\seg_{\kappa,l}$ (Sec. \ref{sec:recovery_complete}), we sample radial functions that lie completely in the cell sequence $\class = \nodecell{\seg_{\kappa,l}}$ and intersect any necessary point constraint. We perform a convex decomposition by intersecting rays with $\proj\class$. We find rays $r_1, ... , r_{\numrays}$, ordered in terms of angle from the start point to the end point of the obstacle, originating from the survey point and intersecting a vertex of $\proj\class$ (Fig. \ref{fig:segment_combo}). For each ray $r_{\kappa,l,k}$, we then find the line segment $z_{\kappa,l,k} = r_{\kappa,l,k} \cap \proj\class$, $k \in \{1, \ldots, \numrays\}$. If the end point is free, then we choose a random point in the final cell of the convex decomposition.

We note that $r_{\kappa,l,k} \cap \proj\class$ will always be a connected set. By virtue of constraint \eqref{eq:constraint_vision}, there will always be a ray that connects adjacent nodes on a survey-cell path since either the nodes will node occlude one another or they will form a psuedolayer.

% todo: pull an example from the supergraph to show how this works
% todo: revisit this

This convex decomposition forms the basis of our parameterization, denoted by $\theta$, of all obstacle curves. For a particular obstacle segment $\obsseg_{\kappa,l}$, we order the line segments $z_{\kappa,l,k} \enspace k \in \{1, \ldots, \numrays\}$, point constraints $x_i$, and join points (if needed) $\joinpt_{\kappa,l-1}$, $\joinpt_{\kappa,l}$ by angle, select a point along each line $z_{\kappa,l,k}$, and connect adjacent points with a line segment. Each point along each line is parameterized by some $\theta_{\kappa,l,k} \in [0,1]$ for finite length $z_{\kappa,l,k}$ and $\theta_{\kappa,l,k} \in [0,\infty)$ for unbounded $z_{\kappa,l,k}$ (Fig. \ref{fig:segment_combo}). %It is clear that the resulting obstacle segment is not self-occluding and lies inside of $\proj\class$.
%The parameterization of these obstacle segments (including free endpoints) is captured in $\mathbf{\theta}$.

To speed the search for consistent environments, we can take advantage of a gradient method based on locally-supporting vertices (obstacle vertices that intersect with the alternative paths). Furthermore, we can also use a gradient-based method to perform ``smoothing" of the obstacles or some other secondary objective. For instance, minimizing the length of the obstacle or minimizing the angles of incidence between two adjacent line segments of the obstacle. See Appendix \ref{app:gradient}.

\section{Incorporating Gradient Ascent in Sampling}\label{app:gradient}

\begin{figure}
    \centering
    \includegraphics[width=0.4\textwidth]{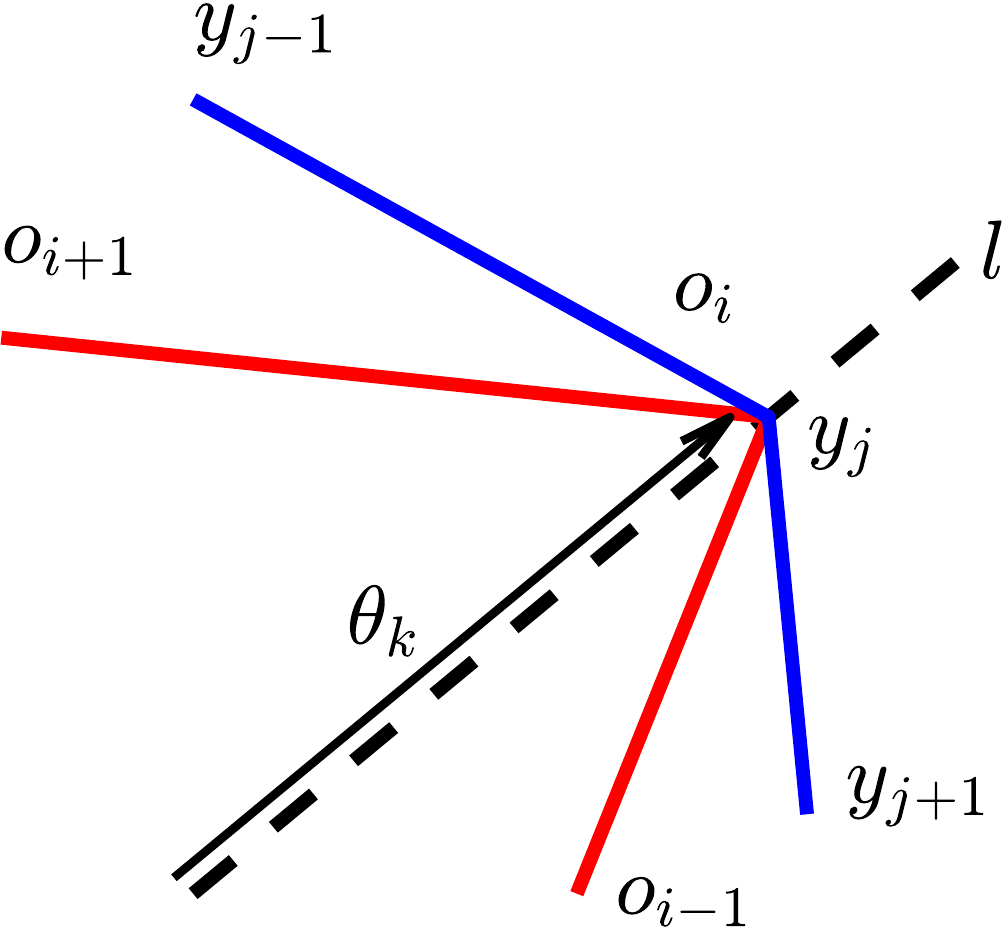}
    \caption{\footnotesize{A visual representation of how the gradient can be calculated. In this case vertex $y_j$ of the robot path is $\epsilon$-close to $o_i$. If we fix the robot path to a particular homotopy, moving $o_i$ along the line $l$ will result in a path where $y_j$ is still locally supported by $o_i$.}}
    \vspace{20pt}
    \label{fig:gradient}
\end{figure}

To verify the obstacles we have sampled are consistent with the demonstration, we want $c(\plan\xi^i_{a_k}) \geq c(\plan\xi^i_{\textrm{inf}})$ for all  $i \in \{1, \ldots, n\}, k \in \{1, \ldots, N^i_a\}$ c.f. (Sec. \ref{sec:verifying_conf}). We define a family of constraints $h_{ik}(O(\mathbf{\theta})) \doteq c(\plan\xi^i_{a_k}) - c(\plan\xi^i_{\textrm{inf}}) \geq 0$ with $\plan\xi^i_{a_k}$ and $\plan\xi^i_{\textrm{inf}}$ planned with respect to $O(\theta)$. We say an obstacle vertex \textit{locally supports} a robot path if and only if the obstacle vertex is $\epsilon$-close to a robot path vertex. For a fixed parameterization $\mathbf{\theta}$, we can calculate the gradient $\frac{\partial h_{ik}}{\partial \mathbf{\theta}}$ using the locally supporting vertices of the obstacle. For a robot path in a fixed homotopy, if the obstacle vertex is parameterized with $\theta_k$, a small enough variation in $\theta_k$ exists such that the same obstacle vertex is locally supporting the robot path. Noting this, the gradient is simply a differentiation of the distance to the two adjacent points on the robot path i.e. $\frac{d}{d\theta_k} \, (|y_j - y_{j-1}| + |y_{j+1} - y_j|)$, summed over all locally supporting points (Fig. \ref{fig:gradient}). The full method is shown in Alg. \ref{alg:m2_find_environment_conf}. In \texttt{CalculateGrad}, we sum all $\frac{\partial h_{ik}}{\partial \mathbf{\theta}}$ if $h_{ij} < 0$. If $h_{ij} \geq 0$, the constraint is satisfied.

\section{Computational Complexity Analysis}\label{app:comp_complex}

The computational complexity of the algorithm is dominated by the size of the CSP. The computational complexity of the CSP is governed by the length of paths found and the branching factor of the survey-cell graph. We now establish the branching factor and path length for the CSP.

\begin{figure}
    \centering
    \includegraphics[width=0.5\textwidth]{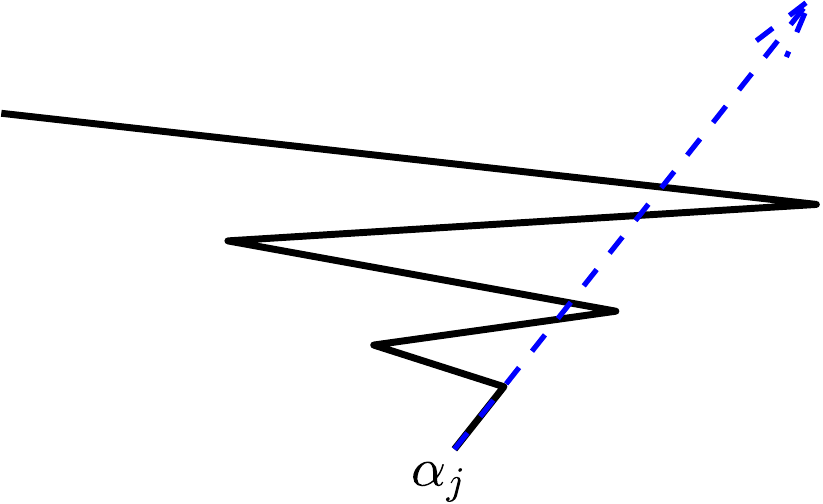}
    \caption{Example of worst case in PCD. The demonstration is in black and a ray of the PCD is in dotted blue. The zig zag structure creates a circumstance where a single ray intersects with $O(N_v)$ lines. However, since the PCD only opens and closes cells where necessary, we only need to open and close cells at the vertex each ray intersects with. Therefore, in the worst case, we open $O(N_v)$ cells.}
    \vspace{20pt}
    \label{fig:pcd_worst_case}
\end{figure}

For the PCD, assume $N_v$ is the number of vertices on both the demonstration and the candidate trajectory. In the worst case this can produce $O(N_v)$ cells, illustrated by Fig. \ref{fig:pcd_worst_case}.

For the survey-cell construction, in the worst case we must consider the intersections of each unique tuple of cells from each PCD. The number of potential tuples we must consider is:

\begin{equation}
    |\Cell_{\spt_1}| \cdot |\Cell_{\spt_2}| \cdot \ldots \cdot |\Cell_{\spt_{N_\spt}}| \approx O(N_v^{N_\spt})
\end{equation}

\noindent since the size of each cell decomposition is $O(N_v)$. A node is created for each cell and survey point so the number of nodes of the resulting graph is $O(N_\spt N_v^{N_\spt})$.

\begin{figure}
    \centering
    \includegraphics[width=0.5\textwidth]{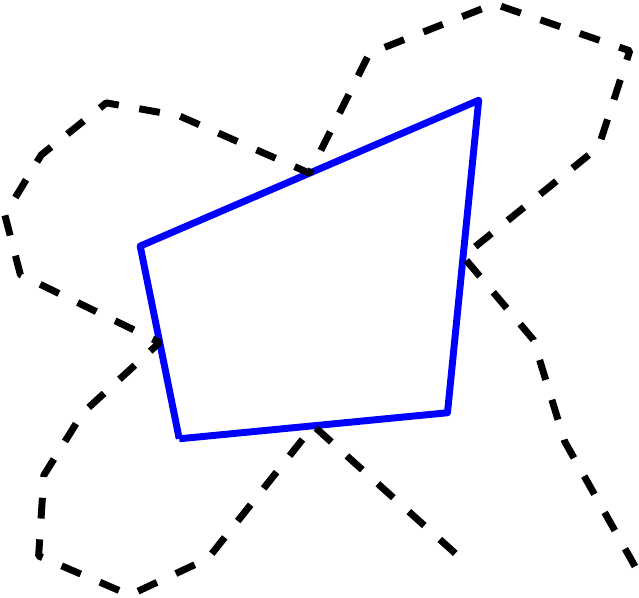}
    \caption{Example of worst case branching factor. A cell in blue has a vertex of the candidate (dashed black) on each of its sides. Each opposing side comes from the PCD of one survey point. Here, there are two survey points (not shown) that result in 4 sides to this cell. Therefore, each side has 2 neighboring cells. So this cell has $4N_\spt$ neighboring cells, or $4N_\spt^2$ neighboring nodes in the graph.}
    \label{fig:branch}
\end{figure}

The branching factor of this graph can be bounded more tightly. If the environment is regular \cite{Lav06}, meaning no three points are exactly collinear, then each side of a cell that does not intersect $\safeset$ can have at most two neighboring cells (see Fig. \ref{fig:branch}). The number of sides is $2N_\spt$, meaning overall there are $4N_\spt$ neighboring cells. Each of the neighboring cells corresponds to $N_\spt$ nodes (one for each survey point) so the branching factor of this graph in the worst case is $O(N_\spt^2)$.

The CSP searches for $n$ paths on this graph. The length of a path is bounded by $O(N_v^{N_\spt})$ since constraint \eqref{eq:shared_cells} prevents any cell from being seen by multiple survey points. The worst case complexity is the branching factor to the power of the length of the path for each of the $n$ paths, or $O(n (N_\spt^2) \, \hat \, (N_v^{N_\spt}) )$.

Class identification is of a similar complexity to the CSP, but instead it only searches over cells instead of survey points and cells (since it ignores vision), so it is of lower complexity. 

Sampling obstacles requires the convex decomposition of a polygon composed of $O(N_v^{N_\spt})$ cells, each with a number of vertices upper bounded by $N_v$. The polygon then has $O(N_v^{N_\spt+1}) = O(N_v^{N_\spt})$ vertices. The convex decomposition can be performed by ordering the rays and performing intersections. Sampling obstacles and searching for a resulting path can be done via $A^*$ and is of much lower complexity that the CSP.

Therefore, the computational complexity is dominated by finding an assignment from the CSP which in the worst case involves evaluating $O(n (N_\spt^2) \, \hat \, (N_v^{N_\spt}) )$ paths.

% todo: improve discussion here

%\section{Special Cases in Obstacle Recovery}\label{app:special_cases_recovery}

%\subsection{Join Point Special Case}

%There is an additional step we must take to ensure that there are no spoked obstacles. If a point constraint exists in either $\seg^j_l(\length{\seg^j_l})$ or $\seg^j_{l+1}(1)$, $1 \leq j \leq m, 1 \leq l \leq \numseg-1$ then we require that this point constraint serves as the join point for the two segments. Otherwise, the resulting segments may produce a spoked obstacle. If this is not possible, in the case that the point constraint does not lie on the shared boundary between the two cells, then we reject this particular assignment as infeasible.

%\subsection{Pseudolayer Special Case}

%Another consideration must be made for paths that switch survey points more than twice in a single psuedolayer from any survey point since these obstacles may self-occlude. In this case we sample the necessary joint points and determine the bounding rays of each segment. Then, each line of convex decomposition is truncated at the bounding rays to ensure there is no parameterization that leads to an obstacle that self-occludes. In some cases, there does not exist a set of join points that result in a parameterization that does not self-occlude. In this case, we reject the assignment as infeasible.

%\input{notation_table.tex}

\end{document}